\documentclass{article}

\usepackage[main, final]{neurips_2026}

\usepackage[utf8]{inputenc} % allow utf-8 input
\usepackage[T1]{fontenc}    % use 8-bit T1 fonts
\usepackage{hyperref}       % hyperlinks
\usepackage{url}            % simple URL typesetting
\usepackage{booktabs}       % professional-quality tables
\usepackage{amsfonts}       % blackboard math symbols
\usepackage{nicefrac}       % compact symbols for 1/2, etc.
\usepackage{microtype}      % microtypography
\usepackage{xcolor}         % colors
\usepackage{xspace}
\usepackage{graphicx}
\usepackage[frozencache,cachedir=_minted]{minted}
\usepackage{amsmath}
\usepackage{booktabs}  
\usepackage{multirow}  
\usepackage{wrapfig}
\usepackage{tcolorbox}
\tcbuselibrary{minted, skins, breakable}

\usepackage{newunicodechar}
\newunicodechar{↔}{\ensuremath{\leftrightarrow}}
\newunicodechar{⇒}{\ensuremath{\Rightarrow}}
\newunicodechar{∀}{\ensuremath{\forall}}
\newunicodechar{∃}{\ensuremath{\exists}}
\newunicodechar{λ}{\ensuremath{\lambda}}
\newunicodechar{≤}{\ensuremath{\leq}}
\newunicodechar{≥}{\ensuremath{\geq}}
\newunicodechar{≠}{\ensuremath{\neq}}
\newunicodechar{∈}{\ensuremath{\in}}
\newunicodechar{∉}{\ensuremath{\notin}}
\newunicodechar{∧}{\ensuremath{\wedge}}
\newunicodechar{∨}{\ensuremath{\vee}}
\newunicodechar{⊢}{\ensuremath{\vdash}}
\newunicodechar{ℕ}{\ensuremath{\mathbb{N}}}
\newunicodechar{ℤ}{\ensuremath{\mathbb{Z}}}
\newunicodechar{ℚ}{\ensuremath{\mathbb{Q}}}
\newunicodechar{ℝ}{\ensuremath{\mathbb{R}}}
\newunicodechar{α}{\ensuremath{\alpha}}
\newunicodechar{β}{\ensuremath{\beta}}
\newunicodechar{γ}{\ensuremath{\gamma}}
\newunicodechar{≡}{\ensuremath{\equiv}}
\newunicodechar{｜}{\textbar}
\newunicodechar{₀}{\ensuremath{_0}}
\newunicodechar{₁}{\ensuremath{_1}}
\newunicodechar{₂}{\ensuremath{_2}}
\newunicodechar{₃}{\ensuremath{_3}}
\newunicodechar{₄}{\ensuremath{_4}}
\newunicodechar{₅}{\ensuremath{_5}}
\newunicodechar{₆}{\ensuremath{_6}}
\newunicodechar{₇}{\ensuremath{_7}}
\newunicodechar{₈}{\ensuremath{_8}}
\newunicodechar{₉}{\ensuremath{_9}}
\newunicodechar{✝}{\ensuremath{\dagger}}
\newunicodechar{∣}{\ensuremath{\mid}}
\newunicodechar{ℂ}{\ensuremath{\mathbb{C}}}
\newunicodechar{▁}{\textunderscore}  % or use \textunderscore
\newunicodechar{ˣ}{$^x$}
\newunicodechar{↦}{$\mapsto$}
\newunicodechar{✓}{\ensuremath{\checkmark}}
\newunicodechar{✗}{\ensuremath{\times}}
\DeclareUnicodeCharacter{2500}{-}     % ─ 替换成 -
\DeclareUnicodeCharacter{2502}{|}     % │ 替换成 |
\DeclareUnicodeCharacter{250C}{+}     % ┌
\DeclareUnicodeCharacter{2510}{+}     % ┐
\DeclareUnicodeCharacter{2514}{+}     % └
\DeclareUnicodeCharacter{2518}{+}     % ┘
\DeclareUnicodeCharacter{251C}{+}     % ├
\DeclareUnicodeCharacter{2524}{+}     % ┤
\DeclareUnicodeCharacter{252C}{+}     % ┬
\DeclareUnicodeCharacter{2534}{+}     % ┴
\DeclareUnicodeCharacter{253C}{+}     % ┼

\definecolor{verylightgray}{gray}{0.95}  % 95% white, 5% black

\newtcolorbox{codepanel}[1][]{
  enhanced, breakable,
  colback=verylightgray, colframe=gray!40,
  boxrule=0.3pt, arc=2pt,
  left=2pt, right=2pt, top=2pt, bottom=2pt,
  #1
}

\newminted[yamlblock]{YAML}{breaklines, fontsize=\scriptsize}
\newminted[yamlblockred]{YAML}{breaklines, fontsize=\scriptsize, bgcolor=diffred}
\newminted[yamlblockgreen]{YAML}{breaklines, fontsize=\scriptsize, bgcolor=diffgreen}

\usepackage[T1]{fontenc}
\usepackage[utf8]{inputenc}
\usepackage{listings}
\usepackage{amssymb}

\usepackage{color}
\definecolor{keywordcolor}{rgb}{0.7, 0.1, 0.1}   % red
\definecolor{tacticcolor}{rgb}{0.0, 0.1, 0.6}    % blue
\definecolor{commentcolor}{rgb}{0.4, 0.4, 0.4}   % grey
\definecolor{symbolcolor}{rgb}{0.0, 0.1, 0.6}    % blue
\definecolor{sortcolor}{rgb}{0.1, 0.5, 0.1}      % green
\definecolor{attributecolor}{rgb}{0.7, 0.1, 0.1} % red

\definecolor{verylightgray}{rgb}{0.97, 0.97, 0.97}
\definecolor{diffred}{rgb}{1.0, 0.85, 0.85}      
\definecolor{diffgreen}{rgb}{0.85, 1.0, 0.85}    
\definecolor{diffreddark}{rgb}{0.6, 0.0, 0.0}    
\definecolor{diffgreendark}{rgb}{0.0, 0.5, 0.0}

\newcommand{\method}{\textbf{{Lean4Agent}}\xspace}
\newcommand{\lib}{\textbf{{FormalAgentLib}}\xspace}
\newcommand{\evolve}{\textbf{{LeanEvolve}}\xspace}
\title{Lean4Agent: Formal Modeling and Verification for Agent Workflow and Trajectory}

\author{
    \textbf{Ruida Wang\textsuperscript{1}}, 
    \textbf{Jerry Huang\textsuperscript{1}}, 
    \textbf{Pengcheng Wang\textsuperscript{1}}, 
    \textbf{Xuanqing Liu\textsuperscript{2}}, 
    \textbf{Luyang Kong\textsuperscript{2}}
    \textbf{Tong Zhang\textsuperscript{1}}
    \\
        \textsuperscript{1}University of Illinois Urbana-Champaign, 
        \textsuperscript{2}Independent researcher
    \\
        \texttt{\{ruidaw,jerry8,pw29\}@illinois.edu} \\
        \texttt{xuanqingliu@outlook.com, luyangkong@gmail.com} \\
        \texttt{tozhang@illinois.edu}
}

\begin{document}

\maketitle

\begin{abstract}
    Equipping Large Language Models (LLMs) to execute reliable multi-step workflows has become a central challenge in artificial intelligence. 
    Despite recent advances in LLMs' agentic capabilities, most agent systems still lack formal methods for specifying, verifying, and debugging their workflow and execution trajectories.
    This challenge mirrors a long-standing problem in mathematics, where the ambiguity of natural languages (NLs) motivates the development of formal languages (FLs). 
    Inspired by this paradigm, we propose \method, to the best of our knowledge, the first framework that uses Lean4, a dependent-type FL to model and verify agent behavior. \method launches \lib, an extensible Lean4 library for formally modeling and verifying agent workflows' semantic consistency under explicit assumptions, and enabling localization of execution-time failures revealed by trajectories.
    Building on \lib, we further develop \evolve, which applies results in \lib to revise workflows to enhance its capability. 
    Extensive experiments on a hard problem subset of SWE-Bench-Verified~\citep{jimenez2024swebench} and a subset of ELAIP-Bench~\citep{dai2025elaipbench} across 5 leading LLMs indicate that the verification-passing workflows outperform the failing ones by an average of \textbf{11.94\%}, and \evolve further improves SWE performance by \textbf{7.47\%} on average. Furthermore, \method establishes a foundation for a new field of using expressive dependent-type FL to formally model and verify agent behavior.
\end{abstract}
\section{Introduction}\label{sec:intro}

Developing Artificial Intelligence (AI) systems with mathematically provable properties has long been a central aspiration of the computer science community~\citep{seshia2022toward}. With the rapid development of LLMs' agentic capabilities, complex LLM-agent workflows are increasingly being deployed in high-stakes domains~\citep{tran2025multi}. This trend heightens the need for formal speculations about LLM agent systems in both workflow specification and execution trajectory levels.

Existing approaches to verifying LLM-based systems remain fragmented in scope and formats. Early approaches, such as LLM-as-judge~\citep{zheng2023judging}, evaluate models' NL outputs but remain vulnerable to hallucination and overconfidence, especially in long-horizon executions~\citep{lin2025llm}. Recent work has introduced formal methods, including simple Hoare-style logic contracts for verifying tool calls~\citep{liu2026toolgate}, SMT-backed verification of action-level policies~\citep{miculicich2025veriguard}, and temporal-logic checking of its artifacts~\citep{ramani2025bridging}.
% and temporal logic model checking of plans translated into Kripke structures~\citep{ramani2025bridging}. 
However, each line of work only addresses part of the problem and is constrained by the expressiveness of its underlying formal language. The temporal logic language is unable to model data-dependent properties, and SMT-based contract checks struggle to express higher-order reasoning. Thus, to the best of our knowledge, existing work does not yet provide a unified framework for formally modeling and verifying agent workflows and trajectories. Both are important for long-horizon autonomous agents~\citep{wang2026agentspex}.

\begin{figure*}[t]
    \vspace{-0.15in}
    \centering
    \includegraphics[width=1.0\columnwidth]{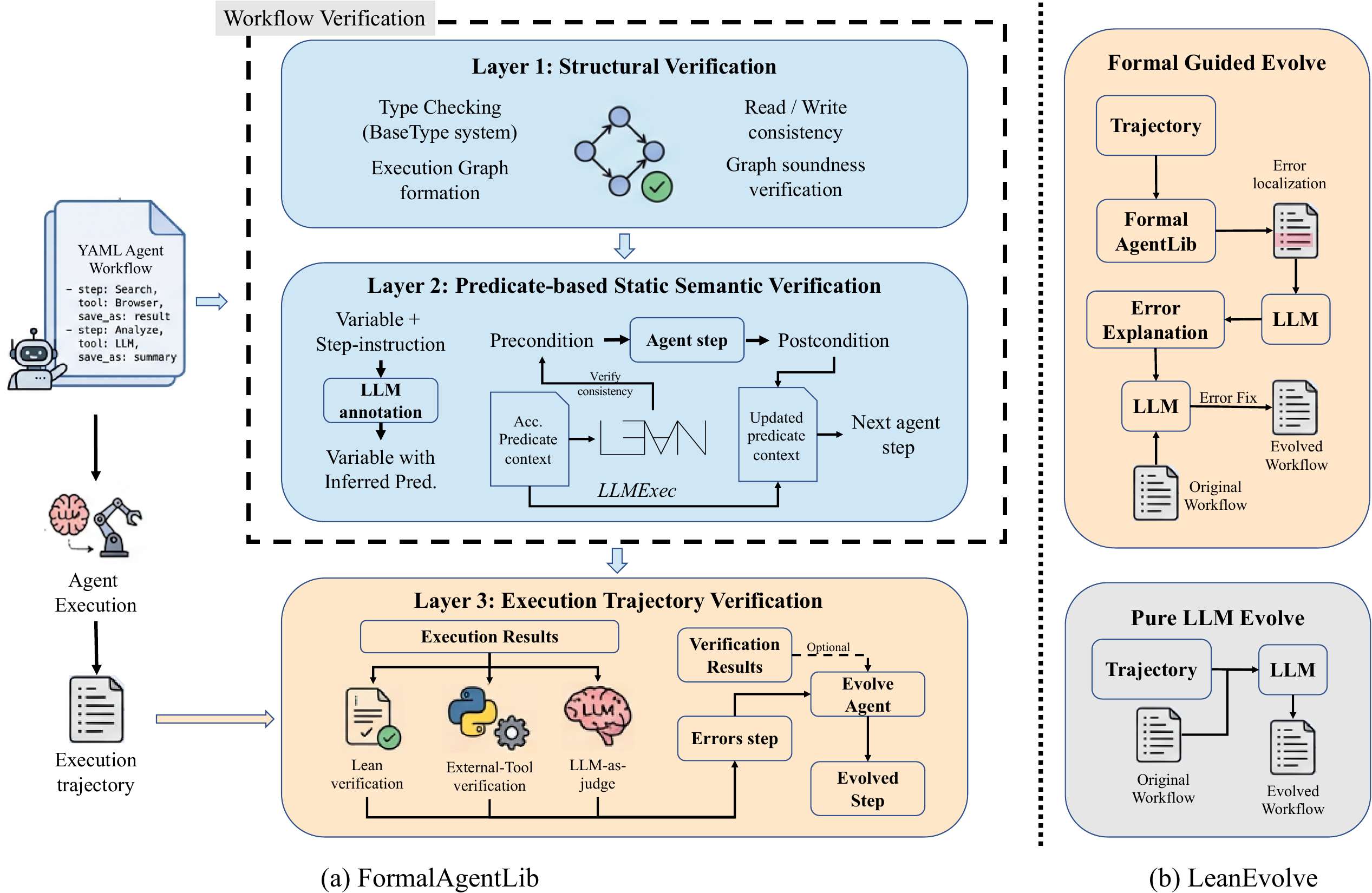}
    \vspace{-0.1in}
    \caption{
    \textbf{\method Framework:} 
    The \method framework consists of two main components. 
    (a) \lib is a three-layer Lean4 library for formally modeling and verifying agent behaviors. 
    Layer 1 verifies the structural correctness of the workflow through a workflow graph. 
    Layer 2 develops a predicate (pred.) system to model pre- and post-conditions of agent executions and verify the semantic self-consistency using \textit{LLMExec} assumptions.
    Layer 3 localizes the violated step in the workflow by applying Lean, external programs, and LLM-as-judge to check the execution trajectory.
    (b) \evolve uses verification results to refine workflows, and it builds formal-guided workflow evolution with the pure-LLM evolve add-on to further enhance its capabilities.
    }
    \label{fig:main}
    \vspace{-0.25in}
\end{figure*}

A parallel challenge has long existed in modern pure mathematics, where most problems involve proving theorems without numerical answers. Because NL is inherently ambiguous, validating complex mathematical arguments becomes increasingly difficult as proofs grow in length and sophistication. To address this issue, mathematicians and computer scientists have adopted dependent type theory~\citep{martin1984intuitionistic} to formally verify proofs.
This paradigm has led to expressive formal languages (FLs) such as Lean~\citep{de2015lean, moura2021lean} and Coq~\citep{coq1996coq}, as well as LLM tools for them. Despite their success in mathematics, the application of FL to uniformly model and verify the LLM agent systems remains largely understudied.

To address these challenges, we propose \method, to the best of our knowledge, the first framework that uses dependent-type FL to uniformly model, verify, and refine agent systems. The overview of \method can be found in Figure~\ref{fig:main}. \method launches \lib, an extensible three-layer Lean4 library for formally modeling and verifying agent workflows and trajectories across three levels of correctness: structural, semantic, and runtime trajectory. Layer 1 verifies the structural well-formedness of the agent workflow, analogous to compiler-level checks for programs. Layer 2 develops a dependent-type predicate system to specify pre- and post-conditions for individual execution steps. It also enables us to uniformly model branching, looping, and submodule composition behaviors. 
With \textit{LLMExec}, under the assumption of LLMs' local correctness, we are able to automate the proof of static semantic soundness while being extensible to new domains. 
It is done by type matching, Hoare-logic reasoning, and auxiliary theorems proved by \lib. 
This layer enables formal verification of workflow specifications under assumptions before deployment, supporting correct-by-construction workflow design~\citep{seshia2022toward}. 
Layer 3 uses a verified workflow with the help of an LLM to inspect execution trajectories, determine whether step-level pre- and post-conditions are satisfied, and localize the execution step responsible for the failure.

Building on \lib, we further propose \evolve, a runtime workflow-refinement method driven by trajectory verification and optional environment feedback. \evolve utilize an LLM together with \lib verification results to identify flaws in current workflows and revise the specifications, thereby improving workflow's performance.

We summarize our contributions of \method as follows:
(1) We launch \lib, to the best of our knowledge, the first extensible Lean4 library for formally modeling and verifying agent workflows' semantic consistency under explicit assumptions, and enabling localization of failures revealed by trajectories. 
(2) Based on \lib, we propose \evolve, a formal-guided workflow evolution method that uses verification feedback and optional environment signals to refine agent workflows. 
(3) We conduct extensive experiments to evaluate \method through software-engineering (SWE) task using a hard-problem-subset of SWE-Bench-Verified~\citep{jimenez2024swebench} and AI-paper understanding tasks using subset of ELAIP-Bench~\citep{dai2025elaipbench} across five leading LLMs. Compared with workflows that fail verification, \lib-verified workflows achieve average improvements of \textbf{14.80\%} on the SWE task and \textbf{9.07\%} on the ELAIP-Bench subset. With \evolve, verified workflows achieve an additional average improvement of \textbf{7.47}\% on the SWE task. This statistically significant improvement demonstrates the usefulness of the \method's workflow verification and the effectiveness of its refinement method.

Broadly speaking, \method provides a solid and unified basis for verifiable LLM-agent systems and opens the door to future directions in 
training and developing self-improving LLM agents. It also offers principles for modeling long-horizon black-box systems. 
To support further development of the field, we open-source the code in \href{https://github.com/RickySkywalker/Lean4Agent}{https://github.com/RickySkywalker/Lean4Agent}.
% To support further development of the field, we will open-source the code in \href{https://github.com/RickySkywalker/Lean4Agent}{https://github.com/RickySkywalker/Lean4Agent} in the near future.
\section{Methodology}\label{sec:meth}

This section introduces the design of the \method framework. The goal is to provide a formal foundation for modeling and verifying agent workflows and trajectories under explicit assumptions and to use the formal guidance to improve workflow design. Section~\ref{meth:prelim} introduces key preliminaries, Section~\ref{meth:lib} describes the design of \lib, and Section~\ref{meth:evolve} presents the \evolve method.

\subsection{Preliminaries}\label{meth:prelim}
We define three central concepts used throughout this paper as follows:

\noindent\textbf{LLM Agent:} Following ReAct~\citep{yao2022react}, we define an LLM agent as the model that can perform the reasoning-acting loop, where the model interleaves internal reasoning with task-specific actions. Reasoning enables the models to formulate, track, and revise their plan, while actions allow them to interact with external tools or information sources.

\noindent\textbf{Agent Workflow:} We define agent workflow as an explicit, structured specification of how an LLM agent approaches a task. Formally, we represent a workflow as a heterogeneous graph $\mathcal{G} :=\langle V, E\rangle$, where $V := \{v_i\}_{i = 1}^n$ is the set of execution nodes and $E \subseteq V \times V$ is the set of transitions. Each node is represented as $v_i := \langle r_i, w_i, t_i, \tau_i \rangle$, where $r_i$, and $w_i$ are the set of variables read and written by the node, $t_i$ is the natural language (NL) instruction, and $\tau_i$ is its execution type. 
In our implementation, we use the YAML workflow format provided by AgentSPEX~\citep{wang2026agentspex} because it offers a clear type system for execution steps and transitions. However, our formulation can be adapted to general workflow specifications with the control-flow graph foundation.

\noindent\textbf{Execution Trajectory:} We define an execution trajectory as the actual rollout produced when an LLM agent runs on a workflow. It can be viewed as a valued path over $\mathcal{G}$, formally, it is $\mathcal{T} := s_0 \rightarrow s_1 \rightarrow \cdots \rightarrow s_l$, where $s_i := \langle pre_i, v_{j_i}, gen_i, pos_i \rangle$ with $pre_i$ and $pos_i$ as pre- and post-execution state, $v_{j_i} \in V$ is the workflow node executed at that transition, and $gen_i$ records the LLM reasoning and tool-call trace for a single ReAct-style step.

\subsection{FormalAgentLib}\label{meth:lib}

We now detail the design of \lib, an extensible Lean4 library for uniformly modeling and verifying LLM-agent workflows and trajectories. To the best of our knowledge, 
it is the first effort to use expressive dependent-type FL to verify agent behaviors.
% it is the first framework that provides such verification in a dependent-type formal language. 
\lib is organized into three complementary layers: structural verification (Section~\ref{lib:struct}), semantic verification (Section~\ref{lib:semantic}), and trajectory-level analysis (Section~\ref{lib:traj}). 

\subsubsection{Layer 1: Structural Verification}\label{lib:struct}

This layer provides the foundation for verifying LLM-independent structural properties of agent workflows. It defines a basic type system for variables, execution nodes, and graph transitions, enabling verification of the workflow's structural well-formedness. Due to the space constraints, the full definitions are presented in Appendix~\ref{app_lib:layer1}.

We construct this layer by first defining data types for workflow variables. Specifically, we introduce the Lean inductive type \texttt{BaseType} and implement compatibility relations following the ideas in~\cite{siek2006gradual, flanagan2006hybrid}. 
We further define \texttt{StepType}, an inductive type for modeling different kinds of execution nodes.
Among these, \texttt{step} and \texttt{task} are most central, where \texttt{step} performs an LLM query with conversation history, whereas \texttt{task} executes without such history. Together, \texttt{BaseType} and \texttt{StepType} define \texttt{WorkflowNode}, the Lean modeling of an agent step. For a node $v_i = \langle r_i, w_i, t_i, \tau_i\rangle$, we represent $r_i = \{r_{i_j}\}_{j = 1}^{|r_i|}, w_i = \{w_{i_j}\}_{j = 1}^{|w_i|}$, where $r_{i_j}, w_{i_j}$ are typed variables and $\tau_i : \texttt{StepType}$.

We next define \texttt{WorkflowEdge} to model node-transition patterns, including sequential execution, branching, and looping. Using these foundation types, we formally model an agent workflow as:
\[
\mathcal{W} = (V, E, v_{entry}, X, P)
\]
where $V$ and $E$ are finite sets of typed nodes and edges, $v_{entry} \in V$ is the entry node, $X \subseteq V$ denotes the exit nodes, and $P$ is the list of typed variables representing initial parameters in the context.

The layer 1 modeling enables structural verification, such as node reachability, edge validity, and read/write consistency, analogous to simple compiler-level verifications for programs. For instance, we can detect read inconsistency when a node reads a variable that is neither an initial parameter nor produced by a reachable predecessor. Appendix~\ref{app_lib:layer1:error} provides a detailed example of such an error.

\subsubsection{Layer 2: Static Semantic Verification for Agent Workflow}\label{lib:semantic}

The second layer models and verifies the static semantic soundness of the agent workflow under explicit assumptions about the local LLM's correctness. To achieve this, we introduce a predicate-based semantic verification system with three components: 
a \textit{Predicate System} that formally models the semantic properties of explicit and implicit variables, 
a \textit{Semantic Workflow Graph} that organizes these predicates across the workflow, 
and a \textit{verification procedure} that verifies whether each step's preconditions are entailed by previously established predicates. Due to the space constraints, we provide the full formal details in Appendix~\ref{app_lib:layer2}.

\paragraph{Predicate System} 
To verify the semantic soundness of a workflow, the first step is to translate the ambiguous NL description into concrete and verifiable constraints. In particular, we define a predicate system on explicit and implicit variables to model such properties

A predicate denotes a decidable property of a variable. We implement the core predicates in Lean as an inductive type, named \texttt{PredicateType} or $\mathcal{P}$. Each base predicate is associated with an interpretation function, \texttt{toProp}, which maps the predicate to a Lean proposition specifying its behaviors. 
For example, \texttt{matchesJsonSchema} takes a variable and a JSON schema, then verifies whether the variable provides a valid JSON value conforming to the given schema. To support task-specific predicates, we introduce \texttt{ext}, which can be associated with a \texttt{PredicateKey} and register user-defined predicates to the predicate universe without modifying the existing definition of \texttt{PredicateType}. For predicates without a clear definition, we model them using the \texttt{custom} predicate, with basic variable-name existence verification and NL explanations for future LLM checks.

Because not all semantic requirements correspond to explicit workflow variables, we also introduce implicit variables and graph-level predicates. For instance, our information-flow predicates track which information each step consumes and produces, while context-managing predicates specify which prior contexts a step may observe. These graph-level constraints help us detect non-trivial errors, especially context-management-related inconsistencies that even humans can miss.

\paragraph{Semantic Workflow Graph} 
The predicate system defines basic semantic constraints, while the semantic workflow graph organizes them across the agent workflow. Its key idea is to constrain each LLM step as a Hoare-style pre- and post-condition pair and then propagate those contracts through formal reasoning over the graph.

We specify the semantics of an agent step using a predicate over its variables, with preconditions describing the properties that must hold before the step, and postconditions stating the properties expected to hold afterward. Formally, we represent the Hoare-style contract of an LLM step as a semantic (workflow) node: 
$s_i = \langle\rho^{(pre)}_i, v_i, \rho_i^{(post)}\rangle$
where $v_i \in V$ is the base node in the workflow graph, $\rho_i^{(pre)}$  is the list of required variable predicates, and $\rho_i^{(post)}$ is the list of variable predicates the step is intended to establish. Beyond sequential execution steps, we also define loop and conditional semantic nodes to capture the contracts of iterative and branching behaviors following the principles in~\cite{pratt1976semantical}.

We then organize these semantic nodes into a semantic (workflow) graph:
\[
\mathcal{S} = \langle \mathcal{G}, s_{para}, S, L, C\rangle
\]
where  $s_{para}$ is a synthetic parameter node whose postconditions record predicates over the input parameters, $S$ is the set of ordinary semantic nodes, $L$ is the set of loop nodes, and $C$ is the set of conditional nodes. The graph $\mathcal{S}$ is linked to the Layer-1 workflow graph by the following well-formedness invariant:
\[
\forall v_i \in \mathcal{G}.V, v_i.\texttt{needSemanticSpec} \Rightarrow \exists s_j \in S,\text{ s.t. } i = j
\]

This representation can also model submodule workflows by treating its postconditions in $s_{para}$ as its input contract and the return nodes' preconditions as its output contract. It allows submodules to be integrated into larger workflows. In practice, we use LLMs to annotate the pre- and post- conditions from NL step instructions.

\paragraph{Verification Procedure}
We verify static semantic soundness under the \textit{LLMExec} assumption: if an LLM step is executed in a predicate context satisfying its preconditions, then the execution can produce a context satisfying its postconditions. Formally
\[
\forall s_i \in \mathcal{S}, \forall \Pi, \Pi \models \rho_i^{(pre)} \Rightarrow \exists \Pi' = LLM(v_i, \Pi) \land \Pi'\models \rho_i^{(post)}
\]
where $\Pi$ denotes the current predicate context and $\Pi'$ denotes the context after executing step $v_i$. This assumption follows the previous analysis that LLMs can typically perform short-horizon tasks well~\citep{lin2025llm, team2026mirothinker} and is also the reason why we can use LLMs to annotate predicates for individual steps. 

Under \textit{LLMExec}, layer-2 verifies the local semantic soundness by ensuring that every node's preconditions are satisfied before execution. For example, explicit variables' predicates can detect when a step refers to a JSON field that no predecessor establishes. Appendix~\ref{app_lib:layer2:explicit_var} provides a concrete case. We further analyze the graph-level predicates in Section~\ref{case:lib}.

Passing Layer-2's verification indicates that the workflow is semantically self-contained with respect to the specified predicates and assumptions. When a workflow fails, this layer identifies which step violates which requirements. But such a failure does not imply that the workflow can never produce a correct answer. Rather, it shows that the specification is not self-contained. In this way, Layer-2 can better support correct-by-construction workflow design.

\subsubsection{Layer 3: Execution Trajectory Verification}\label{lib:traj}
The goal of this layer is to check whether the \textit{LLMExec} assumption holds on particular trajectories. For base predicates $\mathcal{P}$, and their compositions, we can directly evaluate through corresponding Lean propositions. For predicates that may be outside Lean's decidable core, such as the URL connective check, we use external Python validators to add an additional layer of verification. For variables that only have NL definitions, we use LLM-as-judge modules as complementary runtime checkers. The LLM judge can consider comprehensive environment feedback, such as the SWE test-case error messages, to support its assessments. By analyzing the trajectory states against step-level postconditions, Layer-3 localizes the failure-inducing steps, enabling runtime self-checking and formal-guided workflow evolution.

Overall, \lib provides a foundation library for using Lean4 formal language to model and verify agent workflows. It also enables localization of execution-time failures by trajectories.

\subsection{LeanEvolve}\label{meth:evolve}

Building on \lib, we propose \evolve, a dual-mode approach for refining agent workflows using execution trajectories, task outcomes, and verification diagnostics. \evolve is applied when a workflow passes Layer-2 semantic verifications but produces incorrect or uncertain results on a particular execution. In this setting, the goal is to revise the workflow so that future runs avoid the localized failure.

The first mode is \textit{formal-guided evolve}. It uses Layer-3's trajectory analysis from \lib, together with optional environment feedback, to localize the likely failure-inducing step and the violated predicates. We provide an LLM agent with that information to revise the corresponding part of the workflow. If such feedback is unavailable, as in AI paper understanding tasks, the method relies solely on trajectory-level verification. For tasks that permit multiple attempts per instance, \evolve also includes a \textit{pure-LLM evolve} add-on mode. This mode performs a stepwise rewrite of the workflow using the trajectory and any available environmental feedback, without formal guidance. It complements formal-guided evolution by allowing broader exploration. After the above refinement, we rerun the evolved workflow on the same problem to test whether it solves the instance.
\section{Experiment}\label{sec:exp}

We evaluate \method on SWE-Bench-Verified~\citep{jimenez2024swebench} and ELAIP-Bench~\citep{dai2025elaipbench} to assess the effectiveness of \lib verification and \evolve-based workflow refinement. Section~\ref{exp:setup} and~\ref{exp:implementation} describe the experiment setup, Section~\ref{exp:results} presents the main results, and Section~\ref{exp:abl} and \ref{exp:case} conduct ablation studies and case studies. Due to the space constraints, additional experiment results can be found in Appendix~\ref{appendix:add_exp}.

\subsection{Experiment Setup}\label{exp:setup}

\subsubsection{Task and Benchmark}\label{setup:data}

We assess \method across two challenging agentic tasks: software engineering (SWE) and AI paper understanding task. These tasks stress different aspects of agent execution, where SWE typically uses relatively direct workflows but requires substantial self-directed exploration, debugging, and code modification with many implicit constraints. In contrast, paper understanding benefits from fine-grained workflow design that guides the model to systematically read, retrieve evidence, and answer questions. Thus, the two settings respectively represent the LLM-intensive and workflow-intensive agent tasks.

For the SWE task, we use SWE-Bench-Verified~\citep{jimenez2024swebench}, a benchmark for solving real-world GitHub issues. It also offers rich feedback from unit tests for \evolve. Following~\citep{robeyns2025self, xia2025live}, we randomly select a 50-problem hard problem subset from the 500 problems in SWE-Bench-Verified to reduce experiment cost. Those problems would typically take a human engineer more than 1 hour to solve~\citep{jimenez2024swebench}. For the paper understanding task, we use ELAIP-Bench~\citep{dai2025elaipbench}, which is a benchmark of 403 multiple-choice questions derived from recently published AI papers. We randomly select 100 problems from the benchmark for the experiment. 

\subsubsection{Experiment Methods}\label{setup:method}

Because no prior experimental methods directly evaluate the formal modeling and verification of the workflows and trajectories of LLM agents, we design the following two experiments. The first assesses the effectiveness of \lib's semantic verification. We prompt an LLM to independently generate multiple candidate workflows, filter out those that failed the linter verification in workflow language. Then automatically generate the Layer-2 specifications and run verification for those workflows. From this set, we randomly select 3 verification-passing workflows and 3 verification-failing workflows for evaluation. We measure the value of semantic verification by comparing the average task-pass rate across benchmarks between the two groups.

The second experiment we design evaluates Layer-3's trajectory verification and \evolve-based refinement. For SWE, where test results provide rich environment feedback, we enable the pure-LLM evolution add-on in \evolve and report the accumulated solved rate after refinement. For paper understanding, where detailed feedback is unavailable, we compare formal-guided evolution against pure-LLM evolution on a 20-problem subset of trajectories that initially fail, thereby directly evaluating the effect of verification-guided localization.

\subsection{Implementation Details}\label{exp:implementation}

In our implementation, we use Lean v4.20.0. We generate 40 workflows and their Layer-2 verification using Claude-Opus-4.6~\citep{anthropic2026opus46}. To evaluate the cross-model robustness of our framework, we run experiments on 5 leading LLMs, including GPT-5.2~\citep{openai2025gpt52}, GLM-5~\citep{zeng2026glm}, Kimi-K2.5~\citep{team2026kimi}, Qwen-3.5-27B~\citep{qwen3.5}, and Gemma-4-31B-it~\citep{farabet2026gemma4}\footnote{Due to the high cost of Claude-Opus, we are unable to run full experiments on this model; we only evaluate it on the Layer-2 SWE-Bench experiment and report it in Appendix~\ref{appendix:claude}.}.As described in Section~\ref{sec:meth}, we apply AgentSPEX~\citep{wang2026agentspex} as the workflow execution engine, setting the temperature to 1.0 and context length to 163,840.

\subsection{Main Results}\label{exp:results}

\begin{table*}
    \centering
    \caption{Main experimental results of \lib-Layer 2 verification.}
    \label{tab:main}
    \vspace{0.05in}
    \small
    \begin{tabular}{ccccccc}
        \toprule
        \multirow{2}{*}{\textbf{Model}}  & \multicolumn{3}{c}{\textbf{SWE-Bench-Verified}} & \multicolumn{3}{c}{\textbf{ELAIP-Bench}} \\
         & \textbf{Avg. Passed} & \textbf{Avg. Failed} & \textbf{Diff. } & \textbf{Avg. Passed} & \textbf{Avg. Failed} & \textbf{Diff.} \\
        \midrule
        \textit{Large Models} \\
        \textbf{GPT-5.2}        & 62.67\%   & 50.00\%   & 12.67\%           & 32.00\%   & 20.33\%   & 11.67\% \\
        \textbf{GLM-5}          & 58.00\%   & 50.67\%   & 7.33\%            & 37.33\%   & 25.00\%   & 10.83\% \\
        \textbf{Kimi-K2.5}      & 61.33\%   & 46.00\%   & 15.33\%           & 36.00\%   & 28.33\%   & 7.67\% \\
        \midrule
        \textit{Small Models} \\
        \textbf{Gemma-4-31B}    & 60.00\%   & 32.67\%   & 27.33\%           & 35.67\%   & 27.33\%   & 8.34\% \\
        \textbf{Qwen-3.5-27B}   & 49.33\%   & 38.00\%   & 11.33\%           & 42.00\%   & 36.67\%   & 5.33\% \\
        \midrule
        \textbf{Average}        & 58.27\%   & 43.47\%   & \textbf{14.80\%}  & 36.60\%   & 27.53\%   & \textbf{9.07\%} \\
        \bottomrule
    \end{tabular}
    \vspace{-0.1in}
\end{table*}

\begin{table}[t]
  \centering
  \begin{minipage}{0.48\linewidth}
    \centering
    \caption{\evolve enhanced performance of SWE-Bench hard problem subset.}
    \label{tab:evolve_swe}
    \small
    \begin{tabular}{ccc}
        \toprule
        \textbf{Model}          & \textbf{Add. Solved}      & \textbf{Overall} \\
        \midrule
        \textbf{GPT-5.2}        & 8.00\%                    & 70.67\% \\
        \textbf{GLM-5}          & 4.67\%                    & 62.67\% \\
        \textbf{Kimi-K2.5}      & 8.67\%                    & 70.00\% \\
        \textbf{Gemma-4-31B}    & 5.33\%                    & 65.33\% \\
        \textbf{Qwen-3.5-27B}   & 10.67\%                   & 60.00\% \\
        \bottomrule
    \end{tabular}
  \end{minipage}\hfill
  \begin{minipage}{0.48\linewidth}
    \caption{Formal-guided evolve compare with pure-LLM evolve in ELAIP-Bench subset.}
    \label{tab:evolve_elaip}
    \centering
    \small
    \begin{tabular}{ccc}
        \toprule
        \textbf{Model}          & \textbf{Pure-LLM}         & \textbf{Formal-guided} \\
        \midrule
        \textbf{GPT-5.2}        & 18.33\%                   & 25.00\% \\
        \textbf{GLM-5}          & 11.67\%                   & 22.33\% \\
        \textbf{Kimi-K2.5}      & 18.33\%                   & 21.67\% \\
        \textbf{Gemma-4-31B}    & 21.67\%                   & 28.33\% \\
        \textbf{Qwen-3.5-27B}   & 6.67\%                    & 13.33\% \\
        \bottomrule
    \end{tabular}
  \end{minipage}
\vspace{-0.15in}
\end{table}

\subsubsection{\lib Verification Results}\label{results:lib}

Table~\ref{tab:main} compares the performance of workflows that pass Layer-2 verification with those that fail across 5 different models and 2 benchmarks. On the hard subset of SWE-Bench-Verified, passing workflows outperform failing ones by \textbf{14.80\%} on average with a 95\% bootstrap CI of $[10.00\%, 19.60\%]$. On the ELAIP-Bench subset, they achieve an average gain of \textbf{9.07\%} with 95\% CI of $[5.66\%, 13.07\%]$. These statistically significant improvements support the effectiveness of \lib's semantic verification.

On detailed analysis of the results, we observe that the gains are generally larger for smaller models, suggesting that weaker agents are more sensitive to workflow quality. A closer analysis suggests that passing-verification SWE workflows typically include more precise variable specifications, well-structured retry loops, and more reasonable context management. In contrast, failing workflows often contain unsatisfied preconditions, lack valid retry mechanisms, or use context-insensitive execution steps cause break information flow. For paper understanding, we also observed similar patterns, with an additional failure mode: some workflows split unified answer-choice evaluation into separate per-choice nodes with context awareness, which may affect later choices' evaluation by earlier ones, violating \texttt{evaluateChoicesIndependent} predicate. These results and findings highlight the effectiveness of \lib in supporting the correct-by-construction agent workflow design.

\subsubsection{\evolve Enhancement Results}\label{results:evolve}

The results for two settings of evaluation for \evolve are presented in Table~\ref{tab:evolve_swe} and~\ref{tab:evolve_elaip}. The Table~\ref{tab:evolve_swe} shows that full \evolve improves the performance of all evaluated models, adding \textbf{7.47\%} solved instances on average and raising the hard problem subset of SWE-Bench's accuracy rate from 56.93\% to 64.40\%. Further analysis suggests that formal-guided refinement is especially helpful for fixes that require understanding edge cases or cross-file modifications, where localized diagnostics provide fine-grained guidance. Pure-LLM evolution, on the other hand, contributes to broader exploration and is useful for instances that benefit from trial-and-error or non-standard solutions. Table~\ref{tab:evolve_elaip} indicates that formal-guided evolution solves \textbf{7.00\%} more initially failed cases on average than pure-LLM evolution. The additional solved instances suggest that predicate-based diagnostics can provide more fine-grained modification, while pure-LLM revisions typically do not accurately locate the correct step to modify. These findings further support the effectiveness of formal-guided workflow evolution.

\subsection{Ablation Studies}\label{exp:abl}

\subsubsection{Ablation of Graph-level predicates}\label{abl:pred}

We evaluate the contribution of graph-level predicates by removing them from Layer-2 semantic verification and measuring how many originally failing workflows flip to passing. On ELAIP-Bench, 21 out of 40 workflows originally fail the full Layer-2 verification. After removing graph-level predicates, only 8 still fail, indicating that many defects are detected exclusively by graph-level constraints. The most common violation involves \texttt{makeUnifiedJudgement} and \texttt{unifiedLoopBack}, which capture workflow-level consistency requirements that a local predicate alone is unable to model. This result demonstrates the importance of our Layer-2 verification.

\begin{wraptable}{r}{0.35\textwidth} 
  \centering
  \vspace{-0.2in}
  \caption{Enhancement after we drop pure-LLM evolve}
  \label{tab:abl}
  \vspace{0.05in}
  \small
  \begin{tabular}{cc}
    \toprule
    \textbf{Model}          & \textbf{w/o pure-LLM}  \\
    \midrule
    \textbf{GPT-5.2}        & 4.67\% \\
    \textbf{GLM-5}          & 3.33\% \\
    \textbf{Kimi-K2.5}      & 5.33\% \\
    \textbf{Gemma-4-31B}    & 4.00\% \\
    \textbf{Qwen-3.5-27B}   & 8.00\% \\
    \bottomrule
  \end{tabular}
\end{wraptable}

\subsubsection{Ablation of Pure-LLM evolve}\label{abl:llm}

Another experiment we perform is to remove the pure-LLM evolve add-on from the full \evolve on SWE-Bench to assess its contribution. As shown in Table~\ref{tab:abl}, \evolve still improves the performance by 5.07\% on average without this component, but the gain is 2.40\% lower than that of the full system. Inspection of the trajectories suggests that pure-LLM add-on contributes broad exploratory revisions that are not always captured by formal feedback. Nevertheless, the remaining improvements confirm that formal-guided evolution is the primary driver of \evolve's performance.

\subsection{LLM-based predicate annotation study}\label{exp:llm}

This section aims to evaluate whether the LLM-based predicate annotation faithfully represents the intended workflow semantics by manually evaluating all 12 workflows tested in the main experiments. In total, 165 semantic nodes and 1,460 predicate annotations were annotated. We assess both the correctness of the annotated predicates and whether any predicates were omitted. 

Manual evaluation identified 36 mis-specified predicates, corresponding to an annotation error rate of 2.47\%. After manually correcting all 36 annotations, none of the verification results flipped. Among the errors, 34 annotations were overly permissive variable-level conditions, primarily in preconditions. Tightening them did not alter whether the corresponding workflow passed or failed the verification. The remaining 2 annotations were overly weak graph-level predicates annotated in a single verification-failing ELAIP-Bench workflow. It continued to fail after the correction because the failures stemmed from information-flow defects. The results indicate that the local LLM-based annotation and global formal propagation procedure is reliable for the workflow examined in our experiments. It further shows that our framework is robust enough to handle a small fraction of LLM-predicate annotation failures caused by LLM annotators, preventing them from propagating and affecting workflow-level judgment.

\subsection{Case Studies}\label{exp:case}

\subsubsection{Workflow Errors identified by \lib}\label{case:lib}

This case study illustrates how \lib's semantic verification detects non-trivial context management errors before the workflow's roll-out. We present the example YAML workflow in Appendix~\ref{app_case:nontrival}. Appendix~\ref{nontrival:swe} demonstrates a SWE workflow composed entirely of context-isolated \texttt{task} steps. Although the instructions repeatedly refer to information from previous execution turns, the execution type does not present the conversational context. In particular, \texttt{verify\_fix} requires the \texttt{fix\_implementation\_evidence}, whose full details are not available in the context. \lib identifies this information-flow error using the implicit variable's predicate system. Appendix~\ref {nontirval:elaip} presents another non-trivial error in ELAIP-Bench's results. In the workflow, each answer is evaluated by a context-aware \texttt{step}, so later option judgments may be influenced by the earlier ones. This violates the \texttt{evaluateChoicesIndependent} predicate, which requires independent or uniform evaluation of choices. These two cases show that \lib's verification can distinguish between completed context and information management cases. They also demonstrate \lib's potential to support correct-by-construction workflow design.

\subsubsection{Workflow evolve study}\label{case:evolve}

This case study illustrates the error localization and repair capabilities of formal-guided evolution in \evolve. We use \texttt{django\_15098} from SWE-Bench running on the verified plan using GPT-5.2 as a representative example. The base workflow is shown in Appendix~\ref{lean_example:yaml}, and the evolution results are demonstrated in Appendix~\ref{app_case:evolve}. In the original execution, the model identifies only a top-level symptom and fails to trace the root cause through import chains, resulting in an incorrect program fix. The layer-3 trajectory verification successfully localizes the failure to the \texttt{verify\_fix} step, the workflow incorrectly treats the fix as verified even though failing tests remain. Guided by this diagnosis, \evolve revises the step instruction to require tracing the full error chain and validating the fix against the relevant tests, rather than stopping at the first apparent cause. In the next round of execution, the new fix from the revised plan passes. This example shows how trajectory-level verification can pinpoint workflow flaws and guide the targeted repairs.

\section{Related Work}\label{sec:relat}

\subsection{Verification for Agentic System}\label{relat:agent}

As agentic systems mature and are increasingly deployed in high-stakes domains, there is a growing need to model and verify their behavior, especially in long-horizon executions. Existing work on agent verification spans several threads. Among them, agent workflow languages and systems~\citep{langchain2022, langgraph2024, zeng2025adl, vaziri2024pdl, wang2026agentspex} themselves provide a form of structural validation by organizing agent execution into typed or graph-based workflows. But they offer only limited guarantees of semantic correctness, and current advanced LLMs seldom make structural mistakes, making the workflow language linter less useful. Another line of work verifies the tool or tool-generated artifacts~\citep{miculicich2025veriguard, liu2026toolgate, doshi2026towards}, which is crucial for reliable LLM agents but typically targets specific behaviors. To the best of our knowledge, no existing framework formally models and verifies agent workflow and trajectory in a unified manner.

\subsection{LLMs for Formal Theorem Proving}\label{relat:formal}

Using LLMs for formal reasoning has recently become an active research direction. Early work trained neural theorem provers to prove formalized mathematical theorems~\citep{wang2024theoremllama, xin2024deepseek, wang2025ma, ren2025deepseek, lin2025goedel, lin2025goedel2, wang2025gar, chen2025seed}. They establish the basis for modern LLMs' formal reasoning capabilities. Subsequent studies extend formal methods to broader settings, including NL math~\citep{yao2025fans, wang2025let} and physics~\citep{li2025lean4physics, physlib}. Recent work also applies Lean to verify algorithms and programs~\citep{ye2025verina, li2026goedel, zhao2026algoveri}. However, these efforts largely focus on domains with precise specifications and well-defined proof obligations. LLM-agent behavior is different, involving black-box LLM executions, implicit information flows, and trajectory-dependent outcomes. To the best of our knowledge, no prior work has used expressive dependent-type languages to model agent behaviors.
\section{Conclusion}\label{sec:conc}

This paper presents \method, to the best of our knowledge, the first comprehensive framework that applies dependent-type formal language to uniformly model and verify LLM-agent workflow and execution trajectories. \method launches \lib, an extensible Lean4 library for formally modeling and verifying agent workflows' semantic consistency under explicit assumptions. It also enables localization of execution-time failures revealed by trajectories. Based on \lib, we develop \evolve, which refines agent workflows using \lib's verification and optional environment feedback to further improve the task performance. Extensive experiments on SWE and paper understanding tasks across 5 leading models indicate verification-passing workflows outperform failing ones by an average of \textbf{11.94\%}. In addition, workflows refined by \evolve achieve a further average improvement of \textbf{7.47\%} on the SWE task. Furthermore, \method provides a promising foundation for verifiable, self-improving LLM-agent systems through dependent type formal languages.
% Keep this file for further information
\section*{Acknowledgment}
This work was supported in part by NSF grants 2416897, 2505932, and 2601666 and ONR grant N000142512318. Computing resources were provided by the National Center for Supercomputing Applications (NCSA) through Delta (NSF OAC-2005572) and DeltaAI (NSF OAC-2320345), as well as Illinois Computes and the NAIRR Pilot (NAIRR250157).

\newpage
\bibliographystyle{plainnat}
\bibliography{references}

\begin{thebibliography}{49}
\providecommand{\natexlab}[1]{#1}
\providecommand{\url}[1]{\texttt{#1}}
\expandafter\ifx\csname urlstyle\endcsname\relax
  \providecommand{\doi}[1]{doi: #1}\else
  \providecommand{\doi}{doi: \begingroup \urlstyle{rm}\Url}\fi

\bibitem[{Anthropic}(2026)]{anthropic2026opus46}
{Anthropic}.
\newblock Introducing {Claude Opus 4.6}.
\newblock \url{https://www.anthropic.com/news/claude-opus-4-6}, February 2026.
\newblock Accessed: 2026-05-03.

\bibitem[Chase(2022)]{langchain2022}
Harrison Chase.
\newblock Langchain: Building applications with llms through composability.
\newblock \url{https://github.com/langchain-ai/langchain}, 2022.

\bibitem[Chen et~al.(2025)Chen, Chen, Du, Hu, Jiang, Jie, Jin, Jin, Li, Shi, et~al.]{chen2025seed}
Jiangjie Chen, Wenxiang Chen, Jiacheng Du, Jinyi Hu, Zhicheng Jiang, Allan Jie, Xiaoran Jin, Xing Jin, Chenggang Li, Wenlei Shi, et~al.
\newblock Seed-prover 1.5: Mastering undergraduate-level theorem proving via learning from experience.
\newblock \emph{arXiv preprint arXiv:2512.17260}, 2025.

\bibitem[community(2024)]{physlib}
The~Physlib community.
\newblock Physlib: The lean physics library.
\newblock \url{https://github.com/leanprover-community/physlib}, 2024.

\bibitem[Coq(1996)]{coq1996coq}
Projet Coq.
\newblock The coq proof assistant-reference manual.
\newblock \emph{INRIA Rocquencourt and ENS Lyon, version}, 5, 1996.

\bibitem[Dai et~al.(2025)Dai, Hu, Chen, Li, Jin, Zhang, Li, Shang, and Qi]{dai2025elaipbench}
Xinbang Dai, Huikang Hu, Yongrui Chen, Jiaqi Li, Rihui Jin, Yuyang Zhang, Xiaoguang Li, Lifeng Shang, and Guilin Qi.
\newblock Elaipbench: A benchmark for expert-level artificial intelligence paper understanding.
\newblock \emph{arXiv preprint arXiv:2510.10549}, 2025.

\bibitem[De~Moura et~al.(2015)De~Moura, Kong, Avigad, Van~Doorn, and von Raumer]{de2015lean}
Leonardo De~Moura, Soonho Kong, Jeremy Avigad, Floris Van~Doorn, and Jakob von Raumer.
\newblock The lean theorem prover (system description).
\newblock In \emph{Automated Deduction-CADE-25: 25th International Conference on Automated Deduction, Berlin, Germany, August 1-7, 2015, Proceedings 25}, pages 378--388. Springer, 2015.

\bibitem[Doshi et~al.(2026)Doshi, Hong, Xu, Kang, Kapravelos, and K{\"a}stner]{doshi2026towards}
Aarya Doshi, Yining Hong, Congying Xu, Eunsuk Kang, Alexandros Kapravelos, and Christian K{\"a}stner.
\newblock Towards verifiably safe tool use for llm agents.
\newblock \emph{arXiv preprint arXiv:2601.08012}, 2026.

\bibitem[Farabet and Lacombe(2026)]{farabet2026gemma4}
Clement Farabet and Olivier Lacombe.
\newblock Gemma 4: Byte for byte, the most capable open models.
\newblock Google Keyword Blog. \url{https://blog.google/innovation-and-ai/technology/developers-tools/gemma-4/}, April 2026.
\newblock Accessed: 2026-05-03.

\bibitem[Flanagan(2006)]{flanagan2006hybrid}
Cormac Flanagan.
\newblock Hybrid type checking.
\newblock In \emph{Conference record of the 33rd ACM SIGPLAN-SIGACT symposium on Principles of programming languages}, pages 245--256, 2006.

\bibitem[Jimenez et~al.(2024)Jimenez, Yang, Wettig, Yao, Pei, Press, and Narasimhan]{jimenez2024swebench}
Carlos~E Jimenez, John Yang, Alexander Wettig, Shunyu Yao, Kexin Pei, Ofir Press, and Karthik~R Narasimhan.
\newblock {SWE}-bench: Can language models resolve real-world github issues?
\newblock In \emph{The Twelfth International Conference on Learning Representations}, 2024.
\newblock URL \url{https://openreview.net/forum?id=VTF8yNQM66}.

\bibitem[Kwon et~al.(2023)Kwon, Li, Zhuang, Sheng, Zheng, Yu, Gonzalez, Zhang, and Stoica]{kwon2023efficient}
Woosuk Kwon, Zhuohan Li, Siyuan Zhuang, Ying Sheng, Lianmin Zheng, Cody~Hao Yu, Joseph~E. Gonzalez, Hao Zhang, and Ion Stoica.
\newblock Efficient memory management for large language model serving with pagedattention.
\newblock In \emph{Proceedings of the ACM SIGOPS 29th Symposium on Operating Systems Principles}, 2023.

\bibitem[LangChain(2024)]{langgraph2024}
LangChain.
\newblock Langgraph: Build resilient language agents as graphs.
\newblock \url{https://github.com/langchain-ai/langgraph}, 2024.

\bibitem[Li et~al.(2025)Li, Liu, Wang, Ji, He, Pan, Huang, Zhang, and Fung]{li2025lean4physics}
Yuxin Li, Minghao Liu, Ruida Wang, Wenzhao Ji, Zhitao He, Rui Pan, Junming Huang, Tong Zhang, and Yi~R Fung.
\newblock Lean4physics: Comprehensive reasoning framework for college-level physics in lean4.
\newblock \emph{arXiv preprint arXiv:2510.26094}, 2025.

\bibitem[Li et~al.(2026)Li, Yang, Zhao, Zhao, Tang, Yang, Gupta, Su, Jin, et~al.]{li2026goedel}
Zenan Li, Ziran Yang, Haoyu Zhao, Andrew Zhao, Shange Tang, Kaiyu Yang, Aarti Gupta, Zhendong Su, Chi Jin, et~al.
\newblock Goedel-code-prover: Hierarchical proof search for open state-of-the-art code verification.
\newblock \emph{arXiv preprint arXiv:2603.19329}, 2026.

\bibitem[Lin et~al.(2025{\natexlab{a}})Lin, Ning, Zhang, Dong, Liu, Wu, Qi, Sun, Shang, Wang, et~al.]{lin2025llm}
Xixun Lin, Yucheng Ning, Jingwen Zhang, Yan Dong, Yilong Liu, Yongxuan Wu, Xiaohua Qi, Nan Sun, Yanmin Shang, Kun Wang, et~al.
\newblock Llm-based agents suffer from hallucinations: A survey of taxonomy, methods, and directions.
\newblock \emph{arXiv preprint arXiv:2509.18970}, 2025{\natexlab{a}}.

\bibitem[Lin et~al.(2025{\natexlab{b}})Lin, Tang, Lyu, Wu, Lin, Yang, Li, Xia, Chen, Arora, and Jin]{lin2025goedel}
Yong Lin, Shange Tang, Bohan Lyu, Jiayun Wu, Hongzhou Lin, Kaiyu Yang, Jia Li, Mengzhou Xia, Danqi Chen, Sanjeev Arora, and Chi Jin.
\newblock Goedel-prover: A frontier model for open-source automated theorem proving, 2025{\natexlab{b}}.
\newblock URL \url{https://arxiv.org/abs/2502.07640}.

\bibitem[Lin et~al.(2025{\natexlab{c}})Lin, Tang, Lyu, Yang, Chung, Zhao, Jiang, Geng, Ge, Sun, et~al.]{lin2025goedel2}
Yong Lin, Shange Tang, Bohan Lyu, Ziran Yang, Jui-Hui Chung, Haoyu Zhao, Lai Jiang, Yihan Geng, Jiawei Ge, Jingruo Sun, et~al.
\newblock Goedel-prover-v2: Scaling formal theorem proving with scaffolded data synthesis and self-correction.
\newblock \emph{arXiv preprint arXiv:2508.03613}, 2025{\natexlab{c}}.

\bibitem[Liu et~al.(2026)Liu, Peng, Cao, Wang, Deng, Chen, Yin, and Zhang]{liu2026toolgate}
Yanming Liu, Xinyue Peng, Jiannan Cao, Xinyi Wang, Songhang Deng, Jintao Chen, Jianwei Yin, and Xuhong Zhang.
\newblock Toolgate: Contract-grounded and verified tool execution for llms.
\newblock \emph{arXiv preprint arXiv:2601.04688}, 2026.

\bibitem[Martin-L{\"o}f and Sambin(1984)]{martin1984intuitionistic}
Per Martin-L{\"o}f and Giovanni Sambin.
\newblock \emph{Intuitionistic type theory}, volume~9.
\newblock Bibliopolis Naples, 1984.

\bibitem[Miculicich et~al.(2025)Miculicich, Parmar, Palangi, Dvijotham, Montanari, Pfister, and Le]{miculicich2025veriguard}
Lesly Miculicich, Mihir Parmar, Hamid Palangi, Krishnamurthy~Dj Dvijotham, Mirko Montanari, Tomas Pfister, and Long~T Le.
\newblock Veriguard: Enhancing llm agent safety via verified code generation.
\newblock \emph{arXiv preprint arXiv:2510.05156}, 2025.

\bibitem[Moura and Ullrich(2021)]{moura2021lean}
Leonardo~de Moura and Sebastian Ullrich.
\newblock The lean 4 theorem prover and programming language.
\newblock In \emph{Automated Deduction--CADE 28: 28th International Conference on Automated Deduction, Virtual Event, July 12--15, 2021, Proceedings 28}, pages 625--635. Springer, 2021.

\bibitem[{OpenAI}(2025)]{openai2025gpt52}
{OpenAI}.
\newblock Introducing {GPT-5.2}.
\newblock \url{https://openai.com/index/introducing-gpt-5-2/}, December 2025.
\newblock Accessed: 2026-05-03.

\bibitem[{OpenAI}(2026)]{openai2026gpt55}
{OpenAI}.
\newblock Introducing {GPT-5.5}, April 2026.
\newblock URL \url{https://openai.com/index/introducing-gpt-5-5/}.
\newblock Accessed: 2026-05-04.

\bibitem[Pratt(1976)]{pratt1976semantical}
Vaughan~R Pratt.
\newblock Semantical considerations on floyd-hoare logic.
\newblock In \emph{17th Annual Symposium on Foundations of Computer Science (sfcs 1976)}, pages 109--121. IEEE, 1976.

\bibitem[{Qwen Team}(2026)]{qwen3.5}
{Qwen Team}.
\newblock {Qwen3.5}: Towards native multimodal agents, February 2026.
\newblock URL \url{https://qwen.ai/blog?id=qwen3.5}.

\bibitem[Ramani et~al.(2025)Ramani, Tawosi, Alamir, and Borrajo]{ramani2025bridging}
Keshav Ramani, Vali Tawosi, Salwa Alamir, and Daniel Borrajo.
\newblock Bridging llm planning agents and formal methods: A case study in plan verification.
\newblock In \emph{2025 40th IEEE/ACM International Conference on Automated Software Engineering Workshops (ASEW)}, pages 39--42. IEEE, 2025.

\bibitem[Ren et~al.(2025)Ren, Shao, Song, Xin, Wang, Zhao, Zhang, Fu, Zhu, Yang, et~al.]{ren2025deepseek}
ZZ~Ren, Zhihong Shao, Junxiao Song, Huajian Xin, Haocheng Wang, Wanjia Zhao, Liyue Zhang, Zhe Fu, Qihao Zhu, Dejian Yang, et~al.
\newblock Deepseek-prover-v2: Advancing formal mathematical reasoning via reinforcement learning for subgoal decomposition.
\newblock \emph{arXiv preprint arXiv:2504.21801}, 2025.

\bibitem[Robeyns et~al.(2025)Robeyns, Szummer, and Aitchison]{robeyns2025self}
Maxime Robeyns, Martin Szummer, and Laurence Aitchison.
\newblock A self-improving coding agent.
\newblock \emph{arXiv preprint arXiv:2504.15228}, 2025.

\bibitem[Seshia et~al.(2022)Seshia, Sadigh, and Sastry]{seshia2022toward}
Sanjit~A Seshia, Dorsa Sadigh, and S~Shankar Sastry.
\newblock Toward verified artificial intelligence.
\newblock \emph{Communications of the ACM}, 65\penalty0 (7):\penalty0 46--55, 2022.

\bibitem[Siek and Taha(2006)]{siek2006gradual}
Jeremy~G Siek and Walid Taha.
\newblock Gradual typing for functional languages.
\newblock In \emph{Scheme and functional programming workshop}, volume~6, pages 81--92, 2006.

\bibitem[Team et~al.(2026{\natexlab{a}})Team, Bai, Bai, Bao, Cai, Cao, Charles, Che, Chen, Chen, et~al.]{team2026kimi}
Kimi Team, Tongtong Bai, Yifan Bai, Yiping Bao, SH~Cai, Yuan Cao, Y~Charles, HS~Che, Cheng Chen, Guanduo Chen, et~al.
\newblock Kimi k2. 5: Visual agentic intelligence.
\newblock \emph{arXiv preprint arXiv:2602.02276}, 2026{\natexlab{a}}.

\bibitem[Team et~al.(2026{\natexlab{b}})Team, Bai, Bing, Lei, Li, Li, Lin, Min, Su, Wang, et~al.]{team2026mirothinker}
MiroMind Team, S~Bai, L~Bing, L~Lei, R~Li, X~Li, X~Lin, E~Min, L~Su, B~Wang, et~al.
\newblock Mirothinker-1.7 \& h1: Towards heavy-duty research agents via verification.
\newblock \emph{arXiv preprint arXiv:2603.15726}, 2026{\natexlab{b}}.

\bibitem[Tran et~al.(2025)Tran, Dao, Nguyen, Pham, O'Sullivan, and Nguyen]{tran2025multi}
Khanh-Tung Tran, Dung Dao, Minh-Duong Nguyen, Quoc-Viet Pham, Barry O'Sullivan, and Hoang~D Nguyen.
\newblock Multi-agent collaboration mechanisms: A survey of llms.
\newblock \emph{arXiv preprint arXiv:2501.06322}, 2025.

\bibitem[Vaziri et~al.(2024)Vaziri, Mandel, Spiess, and Hirzel]{vaziri2024pdl}
Mandana Vaziri, Louis Mandel, Claudio Spiess, and Martin Hirzel.
\newblock Pdl: a declarative prompt programming language.
\newblock \emph{arXiv preprint arXiv:2410.19135}, 2024.

\bibitem[Wang et~al.(2026)Wang, Huang, Yao, Pan, Niu, Liu, Wang, Lu, Guo, and Zhang]{wang2026agentspex}
Pengcheng Wang, Jerry Huang, Jiarui Yao, Rui Pan, Peizhi Niu, Yaowenqi Liu, Ruida Wang, Renhao Lu, Yuwei Guo, and Tong Zhang.
\newblock Agentspex: An agent specification and execution language.
\newblock \emph{arXiv preprint arXiv:2604.13346}, 2026.

\bibitem[Wang et~al.(2024)Wang, Zhang, Jia, Pan, Diao, Pi, and Zhang]{wang2024theoremllama}
Ruida Wang, Jipeng Zhang, Yizhen Jia, Rui Pan, Shizhe Diao, Renjie Pi, and Tong Zhang.
\newblock Theoremllama: Transforming general-purpose llms into lean4 experts.
\newblock \emph{arXiv preprint arXiv:2407.03203}, 2024.

\bibitem[Wang et~al.(2025{\natexlab{a}})Wang, Li, Fung, and Zhang]{wang2025let}
Ruida Wang, Yuxin Li, Yi~R Fung, and Tong Zhang.
\newblock Let's reason formally: Natural-formal hybrid reasoning enhances llm's math capability.
\newblock \emph{arXiv preprint arXiv:2505.23703}, 2025{\natexlab{a}}.

\bibitem[Wang et~al.(2025{\natexlab{b}})Wang, Pan, Li, Zhang, Jia, Diao, Pi, Hu, and Zhang]{wang2025ma}
Ruida Wang, Rui Pan, Yuxin Li, Jipeng Zhang, Yizhen Jia, Shizhe Diao, Renjie Pi, Junjie Hu, and Tong Zhang.
\newblock Ma-lot: Model-collaboration lean-based long chain-of-thought reasoning enhances formal theorem proving.
\newblock \emph{arXiv preprint arXiv:2503.03205}, 2025{\natexlab{b}}.

\bibitem[Wang et~al.(2025{\natexlab{c}})Wang, Yao, Pan, Diao, and Zhang]{wang2025gar}
Ruida Wang, Jiarui Yao, Rui Pan, Shizhe Diao, and Tong Zhang.
\newblock Gar: Generative adversarial reinforcement learning for formal theorem proving.
\newblock \emph{arXiv preprint arXiv:2510.11769}, 2025{\natexlab{c}}.

\bibitem[Xia et~al.(2025)Xia, Wang, Yang, Wei, and Zhang]{xia2025live}
Chunqiu~Steven Xia, Zhe Wang, Yan Yang, Yuxiang Wei, and Lingming Zhang.
\newblock Live-swe-agent: Can software engineering agents self-evolve on the fly?
\newblock \emph{arXiv preprint arXiv:2511.13646}, 2025.

\bibitem[Xin et~al.(2024)Xin, Ren, Song, Shao, Zhao, Wang, Liu, Zhang, Lu, Du, et~al.]{xin2024deepseek}
Huajian Xin, ZZ~Ren, Junxiao Song, Zhihong Shao, Wanjia Zhao, Haocheng Wang, Bo~Liu, Liyue Zhang, Xuan Lu, Qiushi Du, et~al.
\newblock Deepseek-prover-v1. 5: Harnessing proof assistant feedback for reinforcement learning and monte-carlo tree search.
\newblock \emph{arXiv preprint arXiv:2408.08152}, 2024.

\bibitem[Yao et~al.(2025)Yao, Wang, and Zhang]{yao2025fans}
Jiarui Yao, Ruida Wang, and Tong Zhang.
\newblock Fans--formal answer selection for natural language math reasoning using lean4.
\newblock \emph{arXiv preprint arXiv:2503.03238}, 2025.

\bibitem[Yao et~al.(2022)Yao, Zhao, Yu, Du, Shafran, Narasimhan, and Cao]{yao2022react}
Shunyu Yao, Jeffrey Zhao, Dian Yu, Nan Du, Izhak Shafran, Karthik Narasimhan, and Yuan Cao.
\newblock React: Synergizing reasoning and acting in language models.
\newblock \emph{arXiv preprint arXiv:2210.03629}, 2022.

\bibitem[Ye et~al.(2025)Ye, Yan, He, Kasriel, Yang, and Song]{ye2025verina}
Zhe Ye, Zhengxu Yan, Jingxuan He, Timothe Kasriel, Kaiyu Yang, and Dawn Song.
\newblock Verina: Benchmarking verifiable code generation.
\newblock \emph{arXiv preprint arXiv:2505.23135}, 2025.

\bibitem[Zeng et~al.(2026)Zeng, Lv, Hou, Du, Zheng, Chen, Yin, Ge, Huang, Xie, et~al.]{zeng2026glm}
Aohan Zeng, Xin Lv, Zhenyu Hou, Zhengxiao Du, Qinkai Zheng, Bin Chen, Da~Yin, Chendi Ge, Chenghua Huang, Chengxing Xie, et~al.
\newblock Glm-5: from vibe coding to agentic engineering.
\newblock \emph{arXiv preprint arXiv:2602.15763}, 2026.

\bibitem[Zeng and Yan(2025)]{zeng2025adl}
Sirui Zeng and Xifeng Yan.
\newblock Adl: A declarative language for agent-based chatbots.
\newblock \emph{arXiv preprint arXiv:2504.14787}, 2025.

\bibitem[Zhao et~al.(2026)Zhao, Yang, Li, He, Li, Jin, Veeravalli, Gupta, and Arora]{zhao2026algoveri}
Haoyu Zhao, Ziran Yang, Jiawei Li, Deyuan He, Zenan Li, Chi Jin, Venugopal~V Veeravalli, Aarti Gupta, and Sanjeev Arora.
\newblock Algoveri: An aligned benchmark for verified code generation on classical algorithms.
\newblock \emph{arXiv preprint arXiv:2602.09464}, 2026.

\bibitem[Zheng et~al.(2023)Zheng, Chiang, Sheng, Zhuang, Wu, Zhuang, Lin, Li, Li, Xing, et~al.]{zheng2023judging}
Lianmin Zheng, Wei-Lin Chiang, Ying Sheng, Siyuan Zhuang, Zhanghao Wu, Yonghao Zhuang, Zi~Lin, Zhuohan Li, Dacheng Li, Eric Xing, et~al.
\newblock Judging llm-as-a-judge with mt-bench and chatbot arena.
\newblock \emph{Advances in neural information processing systems}, 36:\penalty0 46595--46623, 2023.

\end{thebibliography}
\newpage
\appendix

\section{Additional Experiment Results}\label{appendix:add_exp}

This section presents additional experimental results omitted from the main paper due to space constraints.

\subsection{Additional results on Claude}\label{appendix:claude}

Due to the high evaluation cost of Claude models, we evaluate Claude 4.5 Opus only on the 50-hard-problem subset of SWE-Bench-Verified~\citep{jimenez2024swebench}. The results indicate workflows that pass verification achieve an average accuracy of 67.33\%, whereas those that fail verification are only 56.67\% on average, yielding a 10.67\% absolute improvement. This result provides additional evidence that the workflows selected by our verification procedure transfer across models.

\subsection{Main results with 95\% CI}\label{add_exp:ci}

\begin{wraptable}{r}{0.4\textwidth} 
  \centering
  \vspace{-0.2in}
  \caption{Full ELAIP-Bench~\cite{dai2025elaipbench} results on Qwen-3.5-27B}
  \label{tab:add_exp_ci_qwen35}
  \vspace{0.05in}
  \small
  \begin{tabular}{cc}
    \toprule
    \textbf{Metrices}           & \textbf{Qwen-3.5-27B}  \\
    \midrule
    \textbf{Avg. Passed}        & 42.10\% \\
    \textbf{Avg. Failed}        & 32.84\% \\
    \textbf{Diff.}              & 9.26\% \\
    \textbf{95\% C.I.}          & [5.79\%, 12.74\%] \\
    \bottomrule
  \end{tabular}
  \vspace{-0.2in}
\end{wraptable}

To assess statistical significance, we report 95\% paired bootstrap confidence intervals (CIs) computed from 10,000 resamples for both benchmarks across the five evaluated models. The intervals are shown in Tables~\ref{tab:CI_SWE} and~\ref{tab:CI_ELAIP}. In nearly all model-task pairs, verification-passing workflows significantly outperform verification-failing ones. The only exception is Qwen-3.5-27B on ELAIP-Bench, where the 95\% CI includes zero; this suggests that the workflow-quality gain is less pronounced for this setting, possibly because the model already achieves comparatively strong baseline performance on the task.

Since the 100-problem ELAIP-Bench subset does not significantly distinguish workflow quality for Qwen-3.5-27B~\cite{qwen3.5}. We conducted an additional evaluation on all 403 problems in ELAIP-Bench. The results are shown in Table~\ref{tab:add_exp_ci_qwen35}. It indicates that when scaling the evaluation to the full dataset, we achieve significant improvement on Qwen-3.5-27B on \method-verified workflows.

\begin{table*}
    \centering
    \caption{SWE-Bench~\cite{jimenez2024swebench} subset's results with 95\% CI}
    \label{tab:CI_SWE}
    \vspace{0.05in}
    \small
    \begin{tabular}{ccccccc}
        \toprule
        \textbf{Model} & \textbf{Avg. Passed} & \textbf{Avg. Failed} & \textbf{Diff.} & \textbf{Lower CI} & \textbf{Upper CI} \\
        \midrule
        \textit{Large Models} \\
        \textbf{GPT-5.2}        & 62.67\%   & 50.00\%   & 12.67\%           & 5.33\%    & 20.67\% \\
        \textbf{GLM-5}          & 58.00\%   & 50.67\%   & 7.33\%            & 0.67\%    & 14.00\% \\
        \textbf{Kimi-K2.5}      & 61.33\%   & 46.00\%   & 15.33\%           & 6.00\%    & 24.67\% \\
        \midrule
        \textit{Small Models} \\
        \textbf{Gemma-4-31B}    & 60.00\%   & 32.67\%   & 27.33\%           & 18.67\%   & 36.00\% \\
        \textbf{Qwen-3.5-27B}   & 49.33\%   & 38.00\%   & 11.33\%           & 2.67\%    & 20.00\% \\
        \midrule
        \textbf{Average}        & 58.27\%   & 43.47\%   & \textbf{14.80\%}  & 10.00\%   & 19.60\% \\
        \bottomrule
    \end{tabular}
    \vspace{-0.1in}
\end{table*}

\begin{table*}
    \centering
    \caption{ELAIP-Bench-Verified~\cite{dai2025elaipbench} subset's results with 95\% CI}
    \label{tab:CI_ELAIP}
    \vspace{0.05in}
    \small
    \begin{tabular}{ccccccc}
        \toprule
        \textbf{Model} & \textbf{Avg. Passed} & \textbf{Avg. Failed} & \textbf{Diff.} & \textbf{Lower CI} & \textbf{Upper CI} \\
        \midrule
        \textit{Large Models} \\
        \textbf{GPT-5.2}        & 32.00\%   & 20.33\%   & 11.67\%           & 5.00\%    & 18.67\% \\
        \textbf{GLM-5}          & 37.33\%   & 25.00\%   & 10.83\%           & 6.33\%    & 18.33\% \\
        \textbf{Kimi-K2.5}      & 36.00\%   & 28.33\%   & 7.67\%            & 1.33\%   & 14.00\% \\
        \midrule
        \textit{Small Models} \\
        \textbf{Gemma-4-31B}    & 35.67\%   & 27.33\%   & 8.34\%            & 2.33\%    & 14.33\% \\
        \textbf{Qwen-3.5-27B}   & 42.00\%   & 36.67\%   & 5.33\%            & -0.33\%   & 10.67\% \\
        \midrule
        \textbf{Average}        & 36.60\%   & 27.53\%   & \textbf{9.07\%}   & 5.66\%    & 13.07\%   \\
        \bottomrule
    \end{tabular}
    \vspace{-0.1in}
\end{table*}

\subsection{LLM-as-judge workflow quality analysis}\label{add_exp:llm}

\begin{table}[t]
  \centering
  \begin{minipage}{0.48\linewidth}
    \centering
    \caption{LLM-as-judge results for SWE-Bench workflows}
    \label{tab:llm_judge_swe}
    \small
    \begin{tabular}{cc}
        \toprule
        \textbf{Model}              & \textbf{LLM score} \\
        \midrule
        \textbf{verified-1}         & 2   \\
        \textbf{verified-2}         & 0   \\
        \textbf{verified-3}         & 8   \\
        \textbf{failed-1}           & 4   \\
        \textbf{failed-2}           & 5  \\
        \textbf{failed-3}           & 8   \\
        \bottomrule
    \end{tabular}
  \end{minipage}\hfill
  \begin{minipage}{0.48\linewidth}
    \caption{LLM-as-judge results for ELAIP-Bench workflows}
    \label{tab:llm_judge_elaip}
    \centering
    \small
    \begin{tabular}{cc}
        \toprule
        \textbf{Model}              & \textbf{LLM score} \\
        \midrule
        \textbf{verified-1}         & 7  \\
        \textbf{verified-2}         & 6   \\
        \textbf{verified-3}         & 2   \\
        \textbf{failed-1}           & 3   \\
        \textbf{failed-2}           & 2  \\
        \textbf{failed-3}           & 2   \\
        \bottomrule
    \end{tabular}
  \end{minipage}
\vspace{-0.15in}
\end{table}

To further evaluate the effectiveness of Layer-2 semantic verification, we compare \lib against the LLM-as-judge baseline using the strong GPT-5.5~\citep{openai2026gpt55}. The judge is prompted with similar workflow descriptions and semantic requirements used for Lean-based verification; the full prompts are provided in \texttt{LLM\_as\_judge\_prompt\_SWE/ELAIP.md}. Results are reported in Tables~\ref{tab:llm_judge_swe} and~\ref{tab:llm_judge_elaip}. On paper-understanding tasks, where workflow quality is relatively easier to assess, the LLM judge largely agrees with \lib’s verification. In contrast, on SWE tasks, whose success criteria are more ambiguous and execution-dependent, LLM-based judgments show weak alignment with both Lean verification and empirical workflow performance. This suggests that pure LLM-based judgment is especially valuable in domains where workflow correctness is difficult to judge from surface-level descriptions alone. A closer inspection reveals that the LLM judge often captures only explicit workflow behavior, while missing implicit information flows, visible context information, and graph-level predicate constraints.

\subsection{Layer-2 guided workflow refinement}\label{add_exp:layer2_guide}

To show that Layer-2 semantic verification can guide workflow repair rather than merely select among existing candidates, we conduct an additional refinement experiment. Starting from the failed-1 workflow analyzed in Appendix~\ref{nontrival:swe}, Layer-2 verification identifies a lack of context continuity across execution steps. We therefore replace the context-isolated \texttt{task} nodes with context-aware \texttt{step} nodes, which makes the workflow pass layer-2 verification. Re-evaluation of the refined workflow using GPT-5.2 on the 50-problem hard subset of SWE-Bench-Verified~\citep{jimenez2024swebench} indicates an accuracy improvement from 52\% to 62\%, yielding a 10\% absolute gain. The same repaired SWE workflow using GLM-5~\citep{zeng2026glm} increased the accuracy from 48.00\% to 60.00\%. 

We also repaired the one failing ELAIP-Bench workflow and evaluated it using GLM-5; the performance improved from 22.00\% to 43.00\%. For the ELAIP-Bench workflow, layer-2 identified the violations of the \texttt{verdictEnumValid} and \texttt{negation\_flag} predicates. The original workflow generated option-level positive or negative judgments, but failed to pass the supporting rationales to the final-answer aggregation step. We repaired the exact information-flow defect by propagating these reasons to the downstream step. After the repair, both workflows reached the performance range of the passing workflows. These controlled before-and-after results provide more distinct evidence that \method does not merely select generally better workflows. Its diagnostics can identify specific defects and guide targeted workflow repair. 

\subsection{Using different model for generating workflow}\label{add_exp:wf_gen}

\begin{wraptable}{r}{0.4\textwidth} 
  \centering
  \vspace{-0.2in}
  \caption{Full ELAIP-Bench~\cite{dai2025elaipbench} results on Qwen-3.5-27B}
  \label{tab:add_exp_diff_mode_gen}
  \vspace{0.05in}
  \small
  \begin{tabular}{cc}
    \toprule
    \textbf{Metrices}           & \textbf{GLM-5}  \\
    \midrule
    \textbf{Avg. Passed}        & 40.67\% \\
    \textbf{Avg. Failed}        & 29.00\% \\
    \textbf{Diff.}              & 11.67\% \\
    \textbf{95\% C.I.}          & [4.33\%, 19.00\%] \\
    \bottomrule
  \end{tabular}
  \vspace{-0.2in}
\end{wraptable}

This experiment aims to assess whether the conclusion we reached in the main experiment, using Opus-4.6 to generate the workflow and annotate layer-2, is general enough to transfer to other models. In this experiment, we generate 40 AI paper understanding workflows using GPT-5.4 under the same workflow-generation instructions and layer-2 annotation pipeline; among them, 24 passed verification and 16 failed, which is similar to our original results. We then randomly sampled 3 passing and 3 failing workflows for verification and evaluated them using GLM-5 under ELAIP-Bench. The result is shown in Table~\ref{tab:add_exp_diff_mode_gen}. We can see that verification-passing workflows significantly outperform the failing workflows. It provides evidence that \method’s capability to distinguish workflow quality is not specific to a single workflow generation and annotation model.

\section{Design Details of Lean4Agent}\label{appendix:design}

This section introduces some important design details and choices for \method.

\subsection{Position of \method compared to existing frameworks}

This section details the difference between \method and prior agent-verification approaches as well as the capabilities uniquely established by the dependent-type (DT) formal language as follows:
\begin{enumerate}
    \item \textbf{Compared with structural verifications:} The most significant difference lies in the rich semantic verification that \method contains through our predicate system. Systems such as LangChain~\citep{langchain2022}, PDL~\citep{vaziri2024pdl}, and AgentSPEX~\citep{wang2026agentspex} primarily support workflow representation, typing, and structural verification. These capabilities are mostly a subset of our layer-1 verification. In our experiments, all evaluated workflows passed the original AgentSPEX linter, yet they exhibited substantial differences in downstream performance. \method can distinguish this difference at layer-2 by verifying that its step-level pre- and post-conditions, information flows, and graph-level predicates are semantically self-consistent.
    \item \textbf{Compared with formal methods for tool-use verification:} The most significant difference is that \method provides a unified and extensible verification framework for both pre-execution workflows and execution trajectories. In particular, ToolGate~\citep{liu2026toolgate} verifies individual tool-mediated state transitions through contracts, while VeriGuard~\citep{miculicich2025veriguard} checks agent actions against safety policies. In contrast, \method models workflow-wide contracts before execution and verifies them at the trajectory level after execution. It therefore provides a unified framework for pre-hoc workflow analysis, runtime failure localization, and verification-guided refinement. Additionally, unlike the fixed set of rules in previous work, we can easily extend to new domains through our extensible predicate system.
    \item \textbf{Advantage of a dependent-type language:} DT is generally a more expressive language compared to traditional SMT-based languages used in agent verification. Lean4, as a DT language, allows specifications to directly represent recursive and nested data structures, higher-order logic relations, and compositional workflow properties. For example, our framework can model recursively nested JSON schemas and specify that every element of a list satisfies a common structured schema. \method can also quantify over semantic environments and abstract black-box agent executions through reusable contracts. Moreover, its extensible predicate system allows domain-specific predicates to be easily implemented and supports judgments without requiring a redesign of the entire verification framework. All the above points are hard to implement in SMT systems.
\end{enumerate}

\subsection{Boundary of the formal components}\label{design:boundary}
Similar to formalized mathematics, \method first abstracts the system's basic semantic requirements and then formally reasons about their relations. We use predicates as the atomic semantic facts of the workflow. Once the workflow structure and predicate annotations are obtained, Lean formally verifies the structural properties, predicate propagation, pre- and post-condition entailment, and workflow-level semantic consistency.

Because manually annotating all workflows is impractical in real-world use cases, we apply LLMs to extract local predicates from step-wise natural language instructions. These predicates are designed to capture a small fraction of facts that can be modeled through program or local LLM judgments. The formal component then composes these local predicates to derive global workflow properties. Therefore, the formal boundary begins after the workflow abstraction and predicate specification. This is analogous to theorem proving in formal mathematics, where the formal part only happens after the statement has been formalized.

\subsection{Types of errors detected by \method framework}\label{design:detect}

We detail the types of errors that can be covered by \method. In particular, \method is an open-ended framework that does not define a fixed and closed list of all possible workflow errors. On the other hand, it detects error families through layer-1's structural verification and extensible predicate verification in layers 2 and 3.

The layer-1 captures structural well-formedness errors, including incompatible variable types, unresolved reads, inconsistent writes, invalid edges, etc. For layer-2 and 3, the semantic predicates in our system can be organized into three categories:

\begin{enumerate}
    \item \textbf{Variable-level predicates:} These predicates detect invalid properties of explicit workflow variables. For example, \texttt{matchJsonSchema} verifies whether the value in the variable satisfies the JSON schema specified in the predicates.
    \item \textbf{Information-level predicates:} These predicates capture whether a step receives and generates the information of the required type for execution, either through explicit variables or the available context. For example, in the SWE task, if a repository exploration step discovers information required by a later testing step but that information is neither stored in a propagated variable nor available in context, the workflow violates this requirement.
    \item \textbf{Graph-level predicates:} These express global behavioral requirements that cannot be determined from a single variable or node. For example, \texttt{verificationCoverage} verifies whether an SWE workflow includes an appropriate stage to validate the fix before submission, and that this validation can span multiple steps.
\end{enumerate}
Due to our highly extensible predicate system, users may register additional domain-specific predicates rather than being restricted to the predicates currently included in our library. Thus, the list is non-exhaustive. 

\subsection{Role of \texttt{LLMExec} assumption}\label{design:llmExec}

The \texttt{LLMExec} assumption is only used in layer-2, where no concrete execution trajectory is yet available. It abstracts the individual LLM step by assuming that, when its preconditions hold, executing the agent can establish the specified postconditions. This abstraction enables static reasoning about whether the workflow is semantically self-contained before deployment. Thus, there is meaning to measure whether the assumption is violated at the layer-2 level, as it is a modeling assumption rather than an empirically evaluated event. 

\subsection{Roles of Lean and LLM-as-judge in layer-3}\label{design:layer3}

In layer 3 of \lib, the Lean4 and external tool verifications directly verify predicates that can be clearly defined in code, such as whether the agent writes specified content to the specified filepath. LLM-as-judge is reserved for predicates whose intended semantics remain primarily in natural language, such as whether a trajectory contains sufficient evidence that the program fix is comprehensive or not. Their failure model includes the following: Formal and external verifications may produce false negatives when the specifications are overly strict or do not fully capture equivalent valid outputs, but their evaluation is deterministic once the predicate is defined. On the other hand, LLM-based judgments can represent more open-ended semantic requirements but may introduce uncertainty. We therefore use them only when the requirement cannot be adequately represented through Lean or an external executable verifier. Across the 1,460 predicate annotations in the evaluated workflows in Section~\ref{exp:llm}, 1,302 are formally or externally checkable, while 158 require LLM-based semantic assessment. However, they have no disagreement problems, since these mechanisms are primarily assigned to different predicates based on their semantics. Therefore, no predicates receive two independent judgments.

\section{Details of \lib implementation}\label{appendix:lib}

The \lib currently contains 151 types and 611 functions to formally model the workflow, along with 41 theorems that help prove properties of agent workflows. The core of \lib is almost entirely written by humans and verified to ensure correctness. We use Lean code to detail the implementation of the type system in this section.

\subsection{Layer 1's details}\label{app_lib:layer1}

\subsubsection{\texttt{BaseType} Full Definition}
The full definition of \texttt{BaseType} is as follows:

\begin{lstlisting}
inductive BaseType where
  | TUnit : BaseType   -- Unit or default type
  | TString : BaseType    -- string value
  | TInt : BaseType   -- integer value
  | TFloat : BaseType -- float value
  | TBool : BaseType  -- boolean value
  | TJson : BaseType  -- unstructured json data value
  | TList : BaseType → BaseType -- list of base types
  | TDict : BaseType -> BaseType -> BaseType  -- dictionary type with key and value types
  | TSet : BaseType → BaseType  -- set of base types
  | TOption : BaseType -> BaseType -- optional base type
  | TRecord : List (String × BaseType) -> BaseType  -- structured record type
  | TUnknown : BaseType  -- unknown type for gradual typing
  deriving Repr, Hashable, Inhabited
\end{lstlisting}
Additionally, we add the proof of DecidableEq for the \texttt{BaseType} to make the proof easier.

\subsubsection{\texttt{StepType} Full Definition}

The full definition of \texttt{StepType} is as follows:

\begin{lstlisting}
inductive StepType where
  -- Pure YAML semantics, no LLM involved.
  | forEachLoop         -- (3) Loop over a list
  | whileLoop           -- (4) Loop with a condition
  | conditional         -- (5) Conditional branch
  | setVariable         -- (7) Set a variable
  | incrementVariable   -- (8) Increment a variable
  | switchBranch        -- (13) Switch/case statement
  | returnValue         -- (16) Return a the value from submodule to parent workflow
  | save                -- (10) Save literal content to a file, expected the file system to change by the LLM
  | input               -- (9) Collect user input → TString
  -- LLM steps with structured output
  | discover            -- (5) Extract structured lists from text output for use in loops and other steps, output can typically be list or json
  | evaluate            -- (11) Quality scoring expected to produce a TFloat
  | validate            -- (12) Criteria check → TBool (PASS/FAIL)
  -- LLM steps with unstructured output
  | task                -- (1) Basic agent action → TString
  | step                -- (2) Agent action with conversation history → TString
  | parallel            -- (15) Parallel execution → TList (module return type)
  -- The composition submodule, which is complicated behavior that cannot be determined until runtime, so we treat it as unstructured.
  | call                -- (14) Call sub-module → depends on module's return type
  | gather              -- (15) Parallel heterogeneous → TList TJson
  | synthesize          -- (6) LLM will create comprehensive outputs and save to files.
  deriving Repr, BEq, Inhabited
\end{lstlisting}

\subsubsection{\texttt{WorkflowNode} Full Definition}

The full definition of \texttt{WorkflowNode} is as follows:

\begin{lstlisting}
structure WorkflowNode where
  /-- Unique numeric identifier -/
  id : NodeId
  /-- Optional human-readable name -/
  name : Option String
  /-- Which of the YAML step types -/
  stepType : StepType
  /-- Variables read from context, with expected types -/
  reads : List TypedVar
  /-- Variables written to context, with their types -/
  writes : List TypedVar
  /-- The instruction/promot string for LLM steps, for op without LLM, this is None. The {{var}} references inside are extracted into `reads`.-/
  llmInstruction : Option String
  deriving Repr, BEq, Inhabited
\end{lstlisting}

\subsubsection{\texttt{WorkflowEdge} Full Definition}

The \texttt{WorkflowEdge}'s type definitions are as follows, where \texttt{NodeId} is the unique identifier of a \texttt{WorkflowNode}, which is the $i$ in $v_i$.
\begin{lstlisting}
inductive WorkflowEdge where
  | seqEdge : (fromNode : NodeId) -> (toNode : NodeId) -> WorkflowEdge -- sequential execution from -> to
  | branchEdge : (cond : NodeId) -> (thenEntry : NodeId) -> (elseEntry : NodeId) -> WorkflowEdge -- conditional
  | loopEdge : (header : NodeId) -> (bodyEntry : NodeId) -> (exit : NodeId) -> WorkflowEdge -- loop edge
  | loopBackEdge : (bodyEnd : NodeId) -> (header : NodeId) -> WorkflowEdge -- when the loop body ends, back to the header to check the loop condition again
  | forkEdge : (forkNode : NodeId) -> (branches : List NodeId) -> WorkflowEdge -- parallel fork: fork node → parallel branch entries
  | joinEdge : (branches : List NodeId) -> (joinNode : NodeId) -> WorkflowEdge -- Parallel join: parallel branch exits → join node
  | switchEdge : (switchNode : NodeId) -> (cases : List NodeId) -> (defaultCase : Option NodeId) -> WorkflowEdge -- Switch: switch node → case entries + optional default
  deriving Repr, BEq, Inhabited
\end{lstlisting}

\subsubsection{\texttt{WorkflowGraph} Full Definition}

The full definition of \texttt{WorkflowGraph} is as follows:

\begin{lstlisting}
structure WorkflowGraph where
  nodes : List WorkflowNode       -- All nodes in the workflow
  edges : List WorkflowEdge       -- All control flow edges
  entry : NodeId                  -- Entry node id (where execution starts)
  exits : List NodeId             -- Exit node ids (may be multiple due to branches)
  parameters : List TypedVar      -- Initial parameters from YAML "parameters" section
\end{lstlisting}

\subsubsection{Error case that layer 1 can identify.}\label{app_lib:layer1:error}

\begin{figure*}[t]
    \vspace{-0.15in}
    \centering
    \includegraphics[width=0.8\columnwidth]{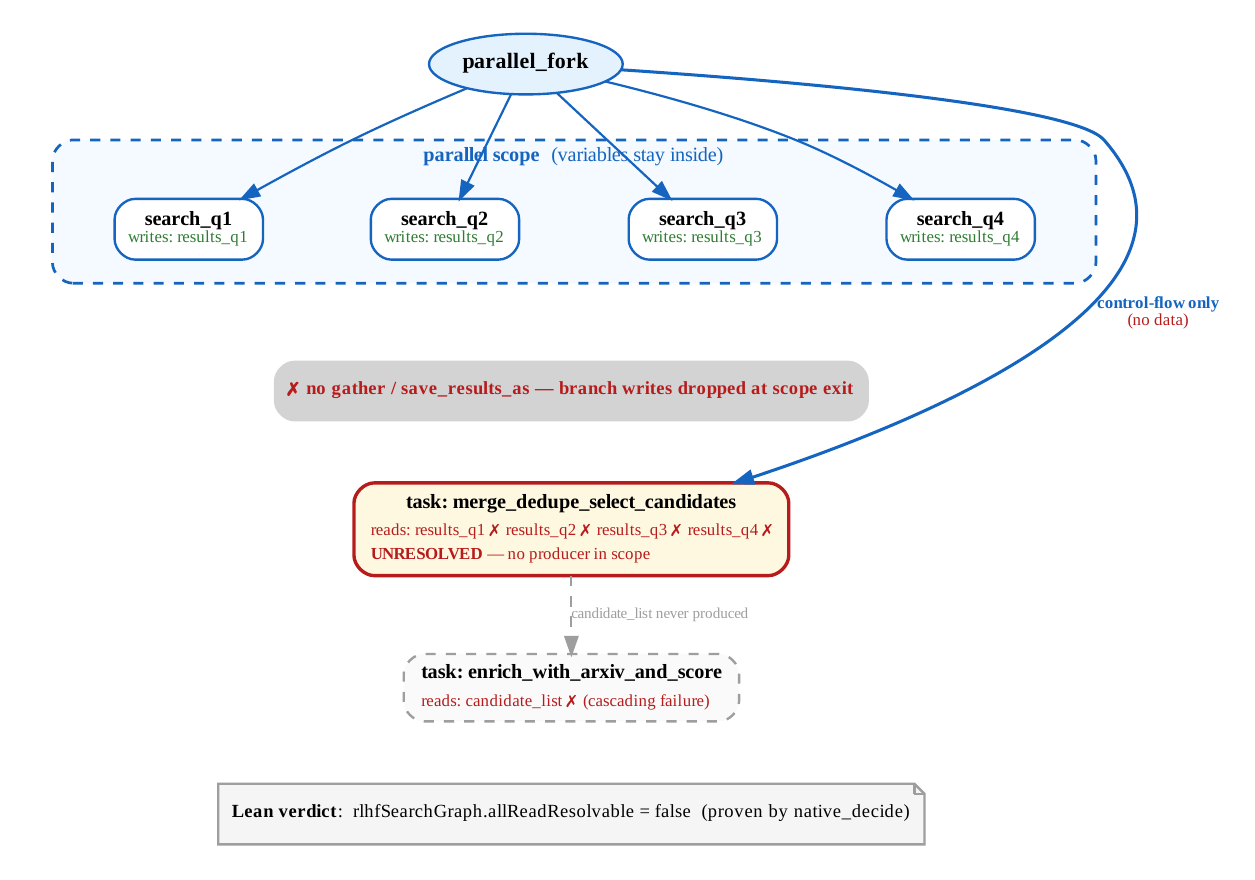}
    \caption{The workflow graph for layer-1 failure}
    \label{fig:layer-1-failure}
    \vspace{-0.2in}
\end{figure*}

Layer-1 verification detects structural errors in the workflow shown in Figure~\ref{fig:layer-1-failure} and~\ref{fig:layer1Error}. In this example, the workflow uses a \texttt{parallel} step to run four LLM branches and produce multiple intermediate results. However, because no \texttt{gather} step propagates these branch-local outputs back to the outer scope, downstream nodes cannot read the variables produced inside the parallel branches. This scoping violation leads to a read-consistency error.

\begin{figure}[H]
\centering
\begin{minted}[breaklines, bgcolor=verylightgray, fontsize=\scriptsize]{YAML}
name: "xxx"
goal: "xxx"
parameters:
  # omitted
config:
  # omitted
workflow:
  - parallel:
    - task:
      # Detail omitted
      - task:
        # Detail omitted
        save_as: "results_q1"
      - task:
        # Detail omitted
        save_as: "results_q2"
      - task:
        # Detail omitted
        save_as: "results_q3"
      - task:
        # Detail omitted
        save_as: "results_q4"
  - task:
      name: "merge_dedupe_select_candidates"
      instruction: |
        You are given four JSON arrays of paper records: {{results_q1}}, {{results_q2}}, {{results_q3}}, {{results_q4}}.
        # Detail omitted
      save_as: "candidate_list"
\end{minted}
\caption{Workflow with layer-1 verified errors}\label{fig:layer1Error}
\end{figure}

\subsection{Layer 2's details}\label{app_lib:layer2}

\subsubsection{\texttt{PredicateType} Full Definition}

In our design of \method, all the predicates are represented through a unified \texttt{PredicateType} framework, including built-in predicates, task-specific extensions, and further extensions. We design this to handle all predicates within the same framework for workflow- and trajectory-level verification.

\begin{lstlisting}
inductive PredicateType where
  | nameExists      -- The base predicate checks whether a variable name exists in the environment.
  | isNonEmptyString -- The base predicate checks whether a variable is a non-empty string.
  | isNonEmptyList   -- The base predicate checks whether a variable is a non-empty list.
  | isValidURL        -- Whether the URL is valid, correcponding to the `isValidURL` predicate in SemanticPredicates
  | isValidFilePath   -- Whether the file path is valid, corresponding to the `isValidFilePath` predicate in SemanticPredicates
  | isValidPath       -- Whether the path (file or URL) is valid, corresponding to the `isValidPath` predicate in SemanticPredicates
  | isValidBibtex   -- Whether the Bibtex entry is valid, corresponding to the `isValidBibtex` predicate in SemanticPredicates
  | isValidLatex   -- Whether the LaTeX string is valid, corresponding to the `isValidLatex` predicate in SemanticPredicates
  | isValidJson   -- Whether the JSON string is valid, corresponding to the `isValidJson` predicate in SemanticPredicates
  | isValidList -- Whether the value is a valid list, corresponding to the `isValidList` predicate in SemanticPredicates
  | toolExists  -- Whether the tool exists in the environment, corresponding to the `toolExists` predicate in SemanticPredicates
  | moduleExists -- Whether the module exists in the environment, corresponding to the `moduleExists` predicate in SemanticPredicates
  | isInt   -- Whether the variable is an integer, corresponding to the `isInt` predicate in temp_SemanticPredicatesExtended
  | matchesJsonSchema (schema : JsonSchema)   -- Whether the provided json schema matches, we perform recursive checking.
  | isJsonWithFields (fields : List JsonFieldSpec)  -- Whether the provided json value has the specified fields, it is sued for compatibility
  | containsSubstring (substring : String)  -- Whether the string variable contains the given substring (for sentinel detection)
  | fileExistsAtPath  -- Whether the variable represents a path to an existing file (runtime check, static just tracks intent)
  | taskCompleted  -- Marker predicate: task has been completed (semantic intent, not runtime verified)
  | custom (name : String) -- The base type for user defined predicates
  | ext (key : PredicateKey) -- The base type for extensible predicates, the semantics of which are defined in the registry
  -- Propositional connectives for composing predicates
  | predicateAnd (p₁ p₂ : PredicateType)  -- Logical AND of two predicates
  | predicateOr (p₁ p₂ : PredicateType)  -- Logical OR of two predicates
  deriving Repr, Inhabited, Hashable
\end{lstlisting}

\subsubsection{\texttt{SemanticWorkflowNode} Full Definition}

\begin{lstlisting}
/-- The semantic environment: a total function from variable names to optional values. -/
def SemanticEnv := String → Option Value

structure SemanticWorkflowNode where
  /-- The Layer 1 node this spec annotates -/
  baseNode : WorkflowNode
  /-- Decidable precondition requirements -/
  precondVariables : List VariablePredicateRequirement := []
  /-- Decidable postcondition facts established -/
  postcondVariables : List VariablePredicateRequirement := []
  /-- Prop precondition (auto-derived) -/
  precond : SemanticEnv → Prop :=
    fun env => ∀ predVar ∈ precondVariables, predVar.toProp env
  /-- Postcondition: After the execution of this node, what should be true in the environment inferred from the LLM prompt or python code.-/
  postcond : SemanticEnv → SemanticEnv → Prop :=
    fun _ env' => ∀ predVar ∈ postcondVariables, predVar.toProp env'
\end{lstlisting}

\subsubsection{\texttt{SemanticWorkflowGraph} Full Definition}

\begin{lstlisting}
structure SemanticWorkflowGraph where
  /-- The underlying Layer 1 WorkflowGraph -/
  baseGraph : WorkflowGraph
  /-- The initial node with the parameter information. Its precondition should be empty and the postcondition should be all requirements on the parameters as well as tools. -/
  paramNode : SemanticWorkflowNode
  /-- Semantic nodes with pre/post conditions -/
  semanticNodes : List SemanticWorkflowNode
  /-- Loop nodes that contains loop-invariants -/
  loopNodes : List SemanticLoopNode := []
  /-- Conditional nodes with branch-aware postconditions -/
  conditionalNodes : List SemanticConditionalNode := []
\end{lstlisting}

\subsubsection{Definition of LLM execution axiom}

\begin{lstlisting}
/-- Axiom: The LLM perform executions just as its instruction specify given the precondition holds, this is the core assumption for static semantic reasoning. -/
axiom llmExecAxiom
  (semanticNode : SemanticWorkflowNode)
  (env : SemanticEnv)
  (precond: semanticNode.precond env) :
  ∃ env', semanticNode.postcond env env'
\end{lstlisting}

\subsubsection{Error case external predicates can locate}\label{app_lib:layer2:explicit_var}

This example shows how external-variable predicates can detect semantic violations in workflow outputs. The erroneous workflow is shown in Figure~\ref{fig:layer2_external_var_error}, and the corresponding Lean verification result is shown in Figure~\ref{fig:layer2_external_var_error_lean}. The step \texttt{compose\_answer} expects fields in \texttt{evidence\_pack} that were never produced by the preceding step: for example, it reads fields such as \texttt{passages} and \texttt{summary}, while the producer emits a different JSON structure. As a result, the required schema predicate is unsatisfied, causing the workflow to fail Lean-based semantic verification.

% This section demonstrates an error identified by a predicate on external variables that violates the requirements. The workflow with error can be found in Figure~\ref{fig:layer2_external_var_error}, and the Lean verification results are in Figure~\ref {fig:layer2_external_var_error_lean}. We can see that the step \texttt{compose\_answer} calls an object in the JSON that does not exist, and this was not present in the previous results. Causing the Lean failure and the workflow is semantically not sound.

\begin{figure}[H]
\centering
\begin{minted}[breaklines, bgcolor=verylightgray, fontsize=\scriptsize]{YAML}
name: "xxx"
goal: "xxx"
parameters:
  # omitted
config:
  # omitted
workflow:
  # ── Node 0: parse_paper ────────────────────────────────────────────────────
  - step:
      name: parse_paper
      instruction: |
        Read the paper below and produce a brief structural overview.
        Identify the title and the main section headings in document order.
        Paper:
        {{paper_content}}
        Return JSON with EXACTLY these keys:
          {
            "title":    <string>,
            "sections": [<string>, ...]
          }
      save_as: paper_data
  # ── Node 1: extract_keywords ───────────────────────────────────────────────
  - step:
      name: extract_keywords
      instruction: |
        Pull 3-7 keywords from the question stem that you would search for
        in the paper.
        Question:
        {{question}}
        Return JSON with EXACTLY these keys:
          {
            "keywords": [<string>, ...]
          }
      save_as: keyword_data
  # ── Node 2: find_evidence ── PRODUCER of the disputed JSON ────────────────
  - step:
      name: find_evidence
      instruction: |
        Using {{keyword_data.keywords}} as search terms, locate verbatim
        passages in the paper that bear on the question.  Each snippet must
        appear word-for-word in the paper.
        Paper:
        {{paper_content}}
        Paper structure: {{paper_data}}
        Return JSON with EXACTLY these keys:
          {
            "snippets": [
              {"text": <string>, "source": <string>},
              ...
            ]
          }
      save_as: evidence_pack
  # ── Node 3: compose_answer ── CONSUMER expecting the WRONG shape ──────────
  - step:
      name: compose_answer
      instruction: |
        Using the evidence packet, write the final answer to the question.
        Cite each passage by quoting it and giving its page number, then
        end with the executive summary supplied alongside the passages.
        Question:
        {{question}}
        Passages:
        {% for p in evidence_pack.passages %}
          - "{{p.quote}}" (page {{p.page}})
        {% endfor %}
        Executive summary:
        {{evidence_pack.summary}}
      save_as: final_answer
\end{minted}
\caption{Workflow with layer-2 verified errors on external variables}\label{fig:layer2_external_var_error}
\end{figure}

\begin{figure}[H]
\centering
\begin{minted}[breaklines, bgcolor=verylightgray, fontsize=\scriptsize]{text}
✗ Node compose_answer (ID 3): missing predicate matchesJsonSchema(AgenticKernel.JsonSchema.jObject
  [("passages",
    AgenticKernel.JsonSchema.jArray
      (AgenticKernel.JsonSchema.jObject
        [("quote", AgenticKernel.JsonSchema.jString), ("page", AgenticKernel.JsonSchema.jNum)])),
   ("summary", AgenticKernel.JsonSchema.jString)]) for variable 'evidence_pack'
  Available predicates (13):
    paper_content: nameExists
    question: nameExists
    paper_content: isNonEmptyString
    question: isNonEmptyString
    paper_data: isNonEmptyString
    paper_data: isValidJson
    paper_data: matchesJsonSchema(AgenticKernel.JsonSchema.jObject
  [("title", AgenticKernel.JsonSchema.jString),
   ("sections", AgenticKernel.JsonSchema.jArray (AgenticKernel.JsonSchema.jString))])
    keyword_data: isNonEmptyString
    keyword_data: isValidJson
    keyword_data: matchesJsonSchema(AgenticKernel.JsonSchema.jObject [("keywords", AgenticKernel.JsonSchema.jArray (AgenticKernel.JsonSchema.jString))])
    evidence_pack: isNonEmptyString
    evidence_pack: isValidJson
    evidence_pack: matchesJsonSchema(AgenticKernel.JsonSchema.jObject
  [("snippets",
    AgenticKernel.JsonSchema.jArray
      (AgenticKernel.JsonSchema.jObject
        [("text", AgenticKernel.JsonSchema.jString), ("source", AgenticKernel.JsonSchema.jString)]))])
\end{minted}
\caption{Lean evaluation results of Fig~\ref{fig:layer2_external_var_error}}\label{fig:layer2_external_var_error_lean}
\end{figure}

\section{Examples for Case Study}\label{appendix:case}

\subsection{Workflow Errors identified by \lib}\label{app_case:nontrival}

\subsubsection{Failed verification in SWE task}\label{nontrival:swe}

The failed verification plan in SWE-Bench:

\begin{minted}[breaklines, bgcolor=verylightgray, fontsize=\scriptsize]{YAML}
name: "swe_agent"
goal: "Given a GitHub issue, reproduce the bug and fix it by modifying source code, then submit a patch"

system_prompt: |
  You are a helpful assistant who can interact multiple times with a computer shell to solve programming tasks.
  STRICT FORMAT RULES:
  - Your response must contain AT MOST one code block.
  - If you include more than one code block, your response will be REJECTED.
  - The code block must contain ONE command (or commands connected with && or ||).
  - The code block must be either a bash block (raw shell command) or a subagent block (subagent name as the fence language).
  - If you have nothing to execute, respond with plain text and NO code block.
  Include a THOUGHT section before your command where you explain your reasoning process.
  <format_example>
  THOUGHT: You should first consider the task goal, your current state, and plan for your future actions. Then tell the user whether you want to run a bash command or call a subagent to help with a task and explain in depth why or why not.

  ```bash
  your_command_here
  ```
  </format_example>

config:
  model: "${MODEL_NAME}"
  temperature: 1.0
  model_kwargs:
    reasoning_effort: "high"
  max_tool_calls_per_step: 20
  expose_submodules_as_tools: false
  enable_inline_tool_calls: true

parameters:
  code_path: "${CODE_PATH:-/workspace/tmp/repo}"
  problem_statement: "${PROBLEM_STATEMENT}"
  regression_test_cmd: ""

workflow:
  # Step 1: Present the problem and orient the agent
  - task:
      name: "setup_and_explore"
      instruction: |
        <pr_description>
        Consider the following PR description:
        {{problem_statement}}
        </pr_description>

        You're a software engineer interacting continuously with a computer by submitting commands.
        You'll be helping implement necessary changes to meet requirements in the PR description.
        Your task is specifically to make changes to non-test files in the current directory in order to fix the issue described in the PR description in a way that is general and consistent with the codebase.
        <IMPORTANT>This is an interactive process where you will think and issue AT LEAST ONE command for every step, see the result, then think and issue your next command(s).</IMPORTANT>

        For each response:
        1. Include a THOUGHT section explaining your reasoning and what you're trying to accomplish
        2. Provide one or more bash tool calls to execute

        IMPORTANT BOUNDARIES:
        - MODIFY: Regular source code files in /testbed (this is the working directory for all your subsequent commands)
        - DO NOT MODIFY: Tests, configuration files (pyproject.toml, setup.cfg, etc.)

        Begin by exploring the repository structure to understand the codebase. Identify the relevant source files that may need to be changed based on the PR description.

        You are operating in an environment where:
        1. You issue at least one command
        2. The system executes the command(s) in a subshell
        3. You see the result(s)
        4. You write your next command(s)
        Each response should include:
        1. **Reasoning text** where you explain your analysis and plan
        2. At least one tool call with your command
        **CRITICAL REQUIREMENTS:**
        - Your response SHOULD include reasoning text explaining what you're doing
        - Your response MUST include AT LEAST ONE bash tool call. You can make MULTIPLE tool calls in a single response when the commands are independent (e.g., searching multiple files, reading different parts of the codebase).
        - Directory or environment variable changes are not persistent. Every action is executed in a new subshell.
        - However, you can prefix any action with `MY_ENV_VAR=MY_VALUE cd /path/to/working/dir && ...` or write/load environment variables from files

        Environment details:
        - You have a full Linux shell environment
        - Always use non-interactive flags (-y, -f) for commands
        - Avoid interactive tools like vi, nano, or any that require user input
        - You can use bash commands or invoke any tool that is available in the environment
        - You can also create new tools or scripts to help you with the task
        - If a tool isn't available, you can also install it

        Start now. Explore the repository at {{code_path}} to understand the project structure, find relevant files, and understand the codebase organization. Use commands like `find`, `ls`, `cat`, `grep` to navigate and understand the code.

  # Step 2: Reproduce the issue
  - task:
      name: "reproduce_issue"
      instruction: |
        <pr_description>
        Consider the following PR description:
        {{problem_statement}}
        </pr_description>

        Based on your exploration of the codebase, now create a script to reproduce the issue described in the PR description. Write a small Python (or appropriate language) script that demonstrates the bug.

        Steps:
        1. Create a reproduction script at /testbed/reproduce_issue.py (or appropriate extension)
        2. Run the script to confirm the issue exists
        3. Analyze the error output to understand the root cause

        IMPORTANT:
        - The working directory for all commands is /testbed
        - Directory or environment variable changes are not persistent
        - Prefix commands with `cd /testbed && ...`
        - Your response MUST include AT LEAST ONE bash tool call

  # Step 3: Identify and implement the fix
  - task:
      name: "implement_fix"
      instruction: |
        <pr_description>
        Consider the following PR description:
        {{problem_statement}}
        </pr_description>

        Now that you've reproduced the issue and understand the root cause, implement a fix.

        Follow this workflow:
        1. Identify the exact source file(s) and location(s) that need to be modified
        2. Understand the surrounding code logic before making changes
        3. Implement the minimal, targeted fix that addresses the issue
        4. Make sure your fix is general and consistent with the existing codebase patterns

        IMPORTANT BOUNDARIES:
        - MODIFY: Regular source code files in /testbed
        - DO NOT MODIFY: Tests, configuration files (pyproject.toml, setup.cfg, etc.)
        - The working directory for all commands is /testbed
        - Directory or environment variable changes are not persistent
        - Prefix commands with `cd /testbed && ...`
        - Use `sed`, `python -c`, or heredoc to edit files (no interactive editors)
        - Your response MUST include AT LEAST ONE bash tool call

  # Step 4: Verify the fix
  - task:
      name: "verify_fix"
      instruction: |
        <pr_description>
        Consider the following PR description:
        {{problem_statement}}
        </pr_description>

        Verify your fix works correctly:
        1. Run your reproduction script again to confirm the issue is resolved
        2. Test edge cases to ensure your fix is robust and doesn't introduce regressions
        3. If a regression test command is available ({{regression_test_cmd}}), run it
        4. If the fix doesn't work, iterate: analyze what went wrong, adjust, and re-test

        IMPORTANT:
        - The working directory for all commands is /testbed
        - Directory or environment variable changes are not persistent
        - Prefix commands with `cd /testbed && ...`
        - Your response MUST include AT LEAST ONE bash tool call

  # Step 5: Submit the patch
  - task:
      name: "submit_patch"
      instruction: |
        Your fix has been verified. Now submit your changes as a git patch.

        Follow these steps IN ORDER, with SEPARATE commands:

        Step 1: Create the patch file
        Run `cd /testbed && git diff -- path/to/file1 path/to/file2 > patch.txt` listing only the source files you modified.
        Do NOT commit your changes.
        <IMPORTANT>
        The patch must only contain changes to the specific source files you modified to fix the issue.
        Do not submit file creations or changes to any of the following files:
        - test and reproduction files (e.g., reproduce_issue.py)
        - helper scripts, tests, or tools that you created
        - installation, build, packaging, configuration, or setup scripts unless they are directly part of the issue you were fixing (you can assume that the environment is already set up for your client)
        - binary or compiled files
        </IMPORTANT>

        Step 2: Verify your patch
        Inspect patch.txt to confirm it only contains your intended changes and headers show `--- a/` and `+++ b/` paths.

        Step 3: Submit (EXACT command required)
        You MUST use this EXACT command to submit:
        ```bash
        echo COMPLETE_TASK_AND_SUBMIT_FINAL_OUTPUT && cat patch.txt
        ```
        If the command fails (nonzero exit status), it will not submit.

        <CRITICAL>
        - Creating/viewing the patch and submitting it MUST be separate commands (not combined with &&).
        - If you modify patch.txt after verifying, you SHOULD verify again before submitting.
        - You CANNOT continue working (reading, editing, testing) in any way on this task after submitting.
        - Unless you think the task is finished and ready for submission, you MUST contain a tool-call in every response. When you make no tool call, it will be treated as a signal that you have already completed the task and submitted your answer, and the system will stop you from making any further tool calls.
        </CRITICAL>
\end{minted}

\subsubsection{Failed verification in ELAIP-Bench}\label{nontirval:elaip}

\begin{minted}[breaklines, bgcolor=verylightgray, fontsize=\scriptsize]{YAML}
name: "elaipbench_agent_bad_plan_4"
goal: "Answer an academic paper question based on the provided passage"
system_prompt: |
  You are an expert academic researcher skilled at reading and understanding scientific papers. You answer questions about academic papers by carefully analyzing the provided passage. Be precise and select only the answer(s) that are supported by the passage.
config:
  model: "${MODEL_NAME}"
  temperature: 1.0
  max_tokens: 163840
  enabled_tools: []
  enable_inline_tool_calls: true
parameters:
  question: ""
  paper_content: ""
  question_type_instruction: ""
  question_type: ""

workflow:
  - step:
      name: skim_paper
      instruction: |
        Read the paper and produce a brief overview.

        Paper:
        {{paper_content}}

        Return JSON:
        {
          "title": "",
          "abstract_summary": ""
        }
      save_as: paper_overview

  - step:
      name: extract_section_headings
      instruction: |
        From the paper below, list the main section headings in the order they appear.

        Paper:
        {{paper_content}}

        Return a JSON array of strings, e.g. ["Introduction", "Method", "Experiments", "Conclusion"].
      save_as: section_headings

  - step:
      name: extract_keywords
      instruction: |
        Read the question stem (do NOT yet look at the four answer options) and extract the most
        discriminative keywords or phrases that should be used to find the relevant evidence in
        the paper.

        Also note explicitly whether the question contains negation cues such as
        "NOT", "incorrect", "wrong", "false", "except", or any similar marker.

        Paper overview:
        {{paper_overview}}
        Section headings: {{section_headings}}
        Question:
        {{question}}

        Return JSON:
        {
          "keywords": [],
          "negation": true or false,
          "stem_summary": ""
        }
      save_as: question_analysis

  - step:
      name: classify_question_type
      instruction: |
        Decide whether the question expects a single correct option (Single-answer) or
        multiple correct options (Multiple-answer).

        Question:
        {{question}}
        {{question_type_instruction}}

        Return one of the strings: "Single-answer" or "Multiple-answer".
      save_as: question_kind

  - step:
      name: search_keywords_in_paper
      instruction: |
        Using the keywords from the question analysis, locate every paragraph in the paper
        that mentions any of those keywords or closely related concepts.

        Keywords: {{question_analysis.keywords}}
        Paper:
        {{paper_content}}

        Return JSON:
        {
          "candidate_paragraphs": [
            {"text": "...", "matched_keywords": []}
          ]
        }
      save_as: candidate_paragraphs

  - step:
      name: filter_relevant_paragraphs
      instruction: |
        From the candidate paragraphs below, keep only those that directly bear on the
        question's stem. Drop tangential matches.

        Candidate paragraphs: {{candidate_paragraphs}}
        Question stem summary: {{question_analysis.stem_summary}}

        Return at most 5 evidence snippets, copied verbatim from the paper.

        Return JSON:
        {
          "evidence_snippets": [
            {"text": "...", "reason": ""}
          ]
        }
      save_as: evidence

  - step:
      name: verify_evidence_quality
      instruction: |
        Briefly judge whether the evidence below is sufficient to discriminate among the
        four answer options. Note any obvious gaps. Do not yet look at the options themselves.

        Question stem: {{question_analysis.stem_summary}}
        Evidence: {{evidence}}

        Return a short paragraph of free-form prose.
      save_as: evidence_check

  - step:
      name: evaluate_A
      instruction: |
        Judge option A independently of any other option.
        Question: {{question}}
        Evidence: {{evidence}}
        Return JSON:
        {
          "verdict": "supported / contradicted / not_established",
          "reason": "",
          "evidence_text": ""
        }
      save_as: judgment_A

  - step:
      name: evaluate_B
      instruction: |
        Judge option B independently of any other option.
        Question: {{question}}
        Evidence: {{evidence}}
        Return JSON:
        {
          "verdict": "supported / contradicted / not_established",
          "reason": "",
          "evidence_text": ""
        }
      save_as: judgment_B

  - step:
      name: evaluate_C
      instruction: |
        Judge option C independently of any other option.
        Question: {{question}}
        Evidence: {{evidence}}
        Return JSON:
        {
          "verdict": "supported / contradicted / not_established",
          "reason": "",
          "evidence_text": ""
        }
      save_as: judgment_C

  - step:
      name: evaluate_D
      instruction: |
        Judge option D independently of any other option.
        Question: {{question}}
        Evidence: {{evidence}}
        Return JSON:
        {
          "verdict": "supported / contradicted / not_established",
          "reason": "",
          "evidence_text": ""
        }
      save_as: judgment_D

  - switch:
      variable: "question_type"
      cases:
        "MA-MCQ":
          - set_variable:
              name: recheck_count
              value: 0

          - while:
              condition: "recheck_count < 3"
              max_iterations: 3
              steps:
                - step:
                    name: aggregate_and_recheck
                    instruction: |
                      The question is multi-answer (typically 2-3 correct options).
                      Per-option judgments so far:
                      A: {{judgment_A}}
                      B: {{judgment_B}}
                      C: {{judgment_C}}
                      D: {{judgment_D}}

                      Count how many options were marked "supported".
                      If fewer than 2 are "supported", revisit the borderline options with
                      a slightly more lenient standard and update their verdicts. Otherwise,
                      keep the verdicts as they are.

                      Return JSON:
                      {
                        "selected_options": [],
                        "selected_count": 0,
                        "notes": ""
                      }
                    save_as: combined_judgment

                - increment: recheck_count

          - step:
              name: finalize_multi_answer
              instruction: |
                Based on the combined judgments below, write the final answer.
                Question: {{question}}
                Combined judgment: {{combined_judgment}}
                {{question_type_instruction}}
              save_as: final_response

        "SA-MCQ":
          - step:
              name: finalize_single_answer
              instruction: |
                Based on the per-option judgments below, select the ONE correct answer.
                Question: {{question}}
                A: {{judgment_A}}
                B: {{judgment_B}}
                C: {{judgment_C}}
                D: {{judgment_D}}
                {{question_type_instruction}}
              save_as: final_response
\end{minted}
\newpage
\subsection{Workflow evolve study example}\label{app_case:evolve}

This is an example of the workflow evolution study. The Lean verification results for the falsified step are shown in Figure~\ref{fig:workflow_evolve_study_example}, and the iterative workflow step is shown in Figure~\ref{fig:case_evolve}.

\begin{figure}[t]
\centering
\begin{minted}[breaklines, bgcolor=verylightgray, fontsize=\scriptsize]{text}
Step 3  (verify_fix)
  trace: 8 LLM iter, 7 tool call(s), tools: [shell_run]

  Layer 2 postcondition contract: 5 predicate(s) (1 runtime + 4 sentinel(s))

    [1] __step_tag_3 : ext(PredicateKey(step_tag, verificatory, []))
      (sentinel — Layer 2 metadata marker; runtime verification not applicable)

    [2] __graph_contrib_3_fix_verified : ext(PredicateKey(graph_level, subGoalContribution, [str(fix_verified)]))
      (sentinel — Layer 2 metadata marker; runtime verification not applicable)

    [3] __graph_verify_3_fix_implemented : ext(PredicateKey(graph_level, subGoalVerification, [str(fix_implemented)]))
      (sentinel — Layer 2 metadata marker; runtime verification not applicable)

    [4] __node_cap_3_implicit_retry_fix_verified : ext(PredicateKey(node_capability, implicitRetry, [str(fix_verified)]))
      (sentinel — Layer 2 metadata marker; runtime verification not applicable)

    [5] fix_verification_evidence : isNonEmptyString
      [Lean]  ✓  typed-state predicate satisfied
      [Tool]  ✗  this predicate claims verification is done but 2 test(s) ultimately failed
      [LLM ]  ✗  Eval evidence shows the instance is unresolved with FAIL_TO_PASS failures in i18n.tests.MiscTests.test_get_language_from_path_real (assertion None != de-1996, etc.) and test_get_supported_language_variant_null, indicating verification did not succeed. The step itself ran the i18n suite (runtests.py i18n --failfast) but tool call #4 returned rc=1, consistent with failing tests.
      ⇒ ✗ FALSIFIED — [Tool] this predicate claims verification is done but 2 test(s) ultimately failed  ||  [LLM]  Eval evidence shows the instance is unresolved with FAIL_TO_PASS failures in i18n.tests.MiscTests.test_get_language_from_path_real (assertion None != de-1996, etc.) and test_get_supported_language_variant_null, indicating verification did not succeed. The step itself ran the i18n suite (runtests.py i18n --failfast) but tool call #4 returned rc=1, consistent with failing tests.

  ⇒ STEP COMPOSITE: ✗ FALSIFIED on predicate(s): fix_verification_evidence
  RE-ROLL: [verify_fix] predicate(s) failed: fix_verification_evidence. Modify the YAML instruction to: replace any keyword-filtered pytest with the explicit FAIL_TO_PASS test path; require zero new failures across the affected test class; do not dismiss collection errors as 'unrelated'.
\end{minted}
\caption{Trajectory verification for workflow evolve example}\label{fig:workflow_evolve_study_example}
\end{figure}

\begin{figure}[H]
\centering
% ---------------- 左栏：Before ----------------
\begin{minipage}[t]{0.48\linewidth}
\centering\small\textbf{Before}\\[2pt]
\begin{codepanel}
\begin{yamlblock}
- step:
  name: "verify_fix"
  instruction: |
    Now verify that your fix resolves the issue.
    Steps:
    1. Re-run the reproduction script you created earlier to confirm the issue is fixed
    2. If there is a regression test command available, run it: {{regression_test_cmd}}
\end{yamlblock}
\begin{yamlblockred}
    2. Run the regression test command: {{regression_test_cmd}}
    3. Examine the test output carefully. If any tests fail, analyze whether they are:
      - FAIL_TO_PASS tests that should now pass with your fix
      - PASS_TO_PASS tests that were passing before but now fail (regression)
    4. If PASS_TO_PASS tests fail, your fix has introduced a regression. A fix that breaks existing functionality is worse than no fix. Revert and refine your approach.
    5. Only proceed when all tests pass
\end{yamlblockred}
\begin{yamlblock}
    Remember:
    - Work in the /testbed directory
    - Every response MUST include at least one bash tool call
\end{yamlblock}
\end{codepanel}
\end{minipage}%
\hfill
% ---------------- 右栏:After ----------------
\begin{minipage}[t]{0.48\linewidth}
\centering\small\textbf{After}\\[2pt]
\begin{codepanel}
\begin{yamlblock}
- step:
  name: "verify_fix"
  instruction: |
    Now verify that your fix resolves the issue.
    Steps:
    1. Re-run the reproduction script you created earlier to confirm the issue is fixed
    2. If there is a regression test command available, run it: {{regression_test_cmd}}
\end{yamlblock}
\begin{yamlblockgreen}
    3. Test edge cases to ensure your fix is robust and doesn't break other functionality
    4. Avoid adding top-level `import` statements to existing modules unless the imported name is unavoidably required at module load time. Prefer function-local imports to sidestep circular-import and import-time-side-effect pitfalls. After any edit that touches imports, sanity-check the module with `python -c 'import <module_path>'` before running tests — if this fails, every downstream test will fail for reasons unrelated to your bug fix.
    5. If anything fails, go back and refine your fix
\end{yamlblockgreen}
\begin{yamlblock}
    Remember:
    - Work in the /testbed directory
    - Every response MUST include at least one bash tool call
\end{yamlblock}
\end{codepanel}
\end{minipage}
\caption{Diff of the \texttt{verify\_fix} step before and after refinement.}
\label{fig:case_evolve}
\end{figure}

\section{Examples of Lean Verification System}\label{appeneix:lean_example}

This section demonstrates one representative workflow in SWE task to demonstrate the Lean verification in detail.

\subsection{Original YAML workflow}\label{lean_example:yaml}

The original YAML workflow for the SWE task is as follows:

\begin{minted}[breaklines, bgcolor=verylightgray, fontsize=\scriptsize]{YAML}
name: "swe_agent"
goal: "Given a GitHub issue, reproduce it and fix it by producing a minimal git patch"

system_prompt: |
    You are a helpful assistant who can interact multiple times with a computer shell to solve programming tasks.
    STRICT FORMAT RULES:
    - Your response must contain AT MOST one code block.
    - If you include more than one code block, your response will be REJECTED.
    - The code block must contain ONE command (or commands connected with && or ||).
    - The code block must be either a bash block (raw shell command) or a subagent block (subagent name as the fence language).
    - If you have nothing to execute, respond with plain text and NO code block.
    Include a THOUGHT section before your command where you explain your reasoning process.
    <format_example>
    THOUGHT: You should first consider the task goal, your current state, and plan for your future actions. Then tell the user whether you want to run a bash command or call a subagent to help with a task and explain in depth why or why not.

config:
    model: "${MODEL_NAME}"
    temperature: 1.0
    model_kwargs:
        reasoning_effort: "high"
    max_tool_calls_per_step: 20
    expose_submodules_as_tools: false
    enable_inline_tool_calls: true

parameters:
    code_path: "${CODE_PATH:-/workspace/tmp/repo}"
    problem_statement: "${PROBLEM_STATEMENT}"
    regression_test_cmd: ""

workflow:
    # ==========================================
    # PHASE 1: Explore and understand the issue
    # ==========================================
    - step:
        name: "explore_repository"
        instruction: |
            <pr_description>
            Consider the following PR description:
            {{problem_statement}}
            </pr_description>

            You're a software engineer interacting continuously with a computer by submitting commands.
            You'll be helping implement necessary changes to meet requirements in the PR description.
            Your task is specifically to make changes to non-test files in the current directory in order to fix the issue described in the PR description in a way that is general and consistent with the codebase.
            <IMPORTANT>This is an interactive process where you will think and issue AT LEAST ONE command for every step, see the result, then think and issue your next command(s).</IMPORTANT>
            For each response:
            1. Include a THOUGHT section explaining your reasoning and what you're trying to accomplish
            2. Provide one or more bash tool calls to execute

            <boundaries>
            - MODIFY: Regular source code files in /testbed (this is the working directory for all your subsequent commands)
            - DO NOT MODIFY: Tests, configuration files (pyproject.toml, setup.cfg, etc.)
            </boundaries>

            <environment_info>
            - You have a full Linux shell environment
            - Always use non-interactive flags (-y, -f) for commands
            - Avoid interactive tools like vi, nano, or any that require user input
            - You can use bash commands or invoke any tool that is available in the environment
            - You can also create new tools or scripts to help you with the task
            - If a tool isn't available, you can also install it
            </environment_info>

            <execution_rules>
            You are operating in an environment where
            1. You issue at least one command
            3. The system executes the command(s) in a subshell
            4. You see the result(s)
            5. You write your next command(s)
            Each response should include:
            1. **Reasoning text** where you explain your analysis and plan
            2. At least one tool call with your command
            **CRITICAL REQUIREMENTS:**
            - Your response SHOULD include reasoning text explaining what you're doing
            - Your response MUST include AT LEAST ONE bash tool call. You can make MULTIPLE tool calls in a single response when the commands are independent (e.g., searching multiple files, reading different parts of the codebase).
            - Directory or environment variable changes are not persistent. Every action is executed in a new subshell.
            - However, you can prefix any action with `MY_ENV_VAR=MY_VALUE cd /path/to/working/dir && ...` or write/load environment variables from files
            Example of a CORRECT response:
            <example_response>
            I need to understand the Builder-related code. Let me find relevant files and check the project structure.
            [Makes multiple bash tool calls: {"command": "ls -la"}, {"command": "find src -name '*.java' | grep -i builder"}, {"command": "cat README.md | head -50"}]
            </example_response>
            </execution_rules>

            Now begin. Start by exploring the repository structure at /testbed to understand the codebase, focusing on files and directories most relevant to the issue described in the PR description.

    # ==========================================
    # PHASE 2: Reproduce the issue
    # ==========================================
    - step:
        name: "reproduce_issue"
        instruction: |
            Now that you've explored the repository, create a script to reproduce the issue described in the PR description.

            Steps:
            1. Based on the PR description and the code you've read, write a small reproduction script (e.g. /testbed/reproduce_issue.py or /testbed/reproduce_issue.sh) that demonstrates the bug or failure.
            2. Run the script and confirm the issue is reproducible. Show the error output.
            3. If the issue is not directly reproducible with a simple script (e.g., it's a behavioral or logic error), explain what you observe and how it differs from expected behavior.

            Remember:
            - Work in the /testbed directory
            - Use non-interactive commands only
            - Every response MUST include at least one bash tool call

    # ==========================================
    # PHASE 3: Locate and fix the source code
    # ==========================================
    - step:
        name: "fix_issue"
        instruction: |
            Now that you've reproduced the issue, locate the relevant source code and implement a fix.

            Steps:
            1. Identify the exact source files and functions that need to be changed
            2. Understand the root cause of the issue by reading the relevant code carefully
            3. Implement a fix that:
               - Addresses the root cause, not just the symptoms
               - Is consistent with the existing codebase style and patterns
               - Is general enough to handle edge cases
               - Does NOT modify any test files or configuration files (pyproject.toml, setup.cfg, etc.)
            4. Use sed, python scripts, or heredocs to make the edits — do NOT use interactive editors

            Remember:
            - Work in the /testbed directory
            - Every response MUST include at least one bash tool call
            - ONLY modify regular source code files

    # ==========================================
    # PHASE 4: Verify the fix
    # ==========================================
    - step:
        name: "verify_fix"
        instruction: |
            Now verify that your fix resolves the issue.

            Steps:
            1. Re-run the reproduction script you created earlier to confirm the issue is fixed
            2. If there is a regression test command available, run it: {{regression_test_cmd}}
            3. Test edge cases to ensure your fix is robust and doesn't break other functionality
            4. If anything fails, go back and refine your fix

            Remember:
            - Work in the /testbed directory
            - Every response MUST include at least one bash tool call

    # ==========================================
    # PHASE 5: Submit the patch
    # ==========================================
    - step:
        name: "create_patch"
        instruction: |
            Your fix is verified. Now create and submit the final patch.

            Follow these steps IN ORDER, with SEPARATE commands:

            Step 1: Create the patch file
            Run `cd /testbed && git diff -- path/to/file1 path/to/file2 > patch.txt` listing only the source files you modified.
            Do NOT commit your changes.
            <IMPORTANT>
            The patch must only contain changes to the specific source files you modified to fix the issue.
            Do not submit file creations or changes to any of the following files:
            - test and reproduction files
            - helper scripts, tests, or tools that you created
            - installation, build, packaging, configuration, or setup scripts unless they are directly part of the issue you were fixing (you can assume that the environment is already set up for your client)
            - binary or compiled files
            </IMPORTANT>

            Step 2: Verify your patch
            Inspect patch.txt to confirm it only contains your intended changes and headers show `--- a/` and `+++ b/` paths.

            Step 3: Submit (EXACT command required)
            You MUST use this EXACT command to submit:
            ```bash
            echo COMPLETE_TASK_AND_SUBMIT_FINAL_OUTPUT && cat patch.txt
            ```
            If the command fails (nonzero exit status), it will not submit.

            <CRITICAL>
            - Creating/viewing the patch and submitting it MUST be separate commands (not combined with &&).
            - If you modify patch.txt after verifying, you SHOULD verify again before submitting.
            - You CANNOT continue working (reading, editing, testing) in any way on this task after submitting.
            - Unless you think the task is finished and ready for submission, you MUST contain a tool-call in every response. When you make no tool call, it will be treated as a signal that you have already completed the task and submitted your answer, and the system will stop you from making any further tool calls.
            </CRITICAL>

\end{minted}

\subsection{Layer-1 and 2 Lean file and its running results}\label{lean–example:layer2}

The transformed Lean file for layer-1 and layer-2 verification is as follows:

\begin{minted}[breaklines, bgcolor=verylightgray, fontsize=\scriptsize]{text}
import Lean
import Mathlib
import AgentVerifier.StaticLayer
import AgentVerifier.StaticSemanticLayer.StaticSemanticLayer
import AgentVerifier.StaticSemanticLayer.WorkflowQualityAnalysis.GraphLevelPredicates

namespace AgenticKernel

-- Node IDs
def good_plan_1_v2_nodeId0 : NodeId := ⟨0⟩
...

-- Node 0: step "explore_repository"
def good_plan_1_v2_node0 : WorkflowNode := {
  id := good_plan_1_v2_nodeId0, name := some "explore_repository"
  stepType := .step
  reads := [⟨"problem_statement", .TString⟩], writes := []
  llmInstruction := some "..."
}

-- Semantic Node 0: step "explore_repository"
def good_plan_1_v2_semNode0 : SemanticWorkflowNode := {
  baseNode := good_plan_1_v2_node0
  precondVariables := [varIsValidTool "shell_run", varIsNonEmptyString "problem_statement"]
  postcondVariables := [markStepExploratory good_plan_1_v2_nodeId0, markSubGoalContribution good_plan_1_v2_nodeId0 "repository_explored", markImplicitRetry good_plan_1_v2_nodeId0 "repository_explored", varIsNonEmptyString "repository_understanding"]
}

-- Node 1: step "reproduce_issue"
def good_plan_1_v2_node1 : WorkflowNode := {
  id := good_plan_1_v2_nodeId1, name := some "reproduce_issue"
  stepType := .step
  reads := [], writes := []
  llmInstruction := some "..."
}

-- Semantic Node 1: step "reproduce_issue"
def good_plan_1_v2_semNode1 : SemanticWorkflowNode := {
  baseNode := good_plan_1_v2_node1
  precondVariables := [varIsValidTool "shell_run", varIsNonEmptyString "repository_understanding"]
  postcondVariables := [markStepTransformative good_plan_1_v2_nodeId1, markSubGoalContribution good_plan_1_v2_nodeId1 "issue_reproduced", markImplicitRetry good_plan_1_v2_nodeId1 "issue_reproduced", varIsNonEmptyString "reproduction_evidence"]
}

-- Node 2: step "fix_issue"
def good_plan_1_v2_node2 : WorkflowNode := {
  id := good_plan_1_v2_nodeId2, name := some "fix_issue"
  stepType := .step
  reads := [], writes := []
  llmInstruction := some "..."
}

-- Semantic Node 2: step "fix_issue"
def good_plan_1_v2_semNode2 : SemanticWorkflowNode := {
  baseNode := good_plan_1_v2_node2
  precondVariables := [varIsValidTool "shell_run", varIsNonEmptyString "reproduction_evidence", varIsNonEmptyString "repository_understanding"]
  postcondVariables := [markStepTransformative good_plan_1_v2_nodeId2, markSubGoalContribution good_plan_1_v2_nodeId2 "fix_implemented", markImplicitRetry good_plan_1_v2_nodeId2 "fix_implemented", varIsNonEmptyString "fix_implementation_evidence"]
}

-- Node 3: step "verify_fix"
def good_plan_1_v2_node3 : WorkflowNode := {
  id := good_plan_1_v2_nodeId3, name := some "verify_fix"
  stepType := .step
  reads := [⟨"regression_test_cmd", .TString⟩], writes := []
  llmInstruction := some "..."
}

-- Semantic Node 3: step "verify_fix"
def good_plan_1_v2_semNode3 : SemanticWorkflowNode := {
  baseNode := good_plan_1_v2_node3
  precondVariables := [varIsValidTool "shell_run", varNameExists "regression_test_cmd", varIsNonEmptyString "fix_implementation_evidence", varIsNonEmptyString "reproduction_evidence"]
  postcondVariables := [markStepVerificatory good_plan_1_v2_nodeId3, markSubGoalContribution good_plan_1_v2_nodeId3 "fix_verified", markSubGoalVerification good_plan_1_v2_nodeId3 "fix_implemented", markImplicitRetry good_plan_1_v2_nodeId3 "fix_verified", varIsNonEmptyString "fix_verification_evidence"]
}

-- Node 4: step "create_patch"
def good_plan_1_v2_node4 : WorkflowNode := {
  id := good_plan_1_v2_nodeId4, name := some "create_patch"
  stepType := .step
  reads := [], writes := []
  llmInstruction := some "..."
}

-- Semantic Node 4: step "create_patch"
def good_plan_1_v2_semNode4 : SemanticWorkflowNode := {
  baseNode := good_plan_1_v2_node4
  precondVariables := [varIsValidTool "shell_run", varIsNonEmptyString "fix_implementation_evidence"]
  postcondVariables := [markStepTransformative good_plan_1_v2_nodeId4, markSubGoalContribution good_plan_1_v2_nodeId4 "patch_submitted", markImplicitRetry good_plan_1_v2_nodeId4 "patch_submitted", varIsNonEmptyString "patch_submission_evidence"]
}

def good_plan_1_v2Graph : WorkflowGraph := {
  nodes := [good_plan_1_v2_node0, good_plan_1_v2_node1, good_plan_1_v2_node2, good_plan_1_v2_node3, good_plan_1_v2_node4]
  edges := [
    .seqEdge good_plan_1_v2_nodeId0 good_plan_1_v2_nodeId1,
    .seqEdge good_plan_1_v2_nodeId1 good_plan_1_v2_nodeId2,
    .seqEdge good_plan_1_v2_nodeId2 good_plan_1_v2_nodeId3,
    .seqEdge good_plan_1_v2_nodeId3 good_plan_1_v2_nodeId4
  ]
  entry := good_plan_1_v2_nodeId0
  exits := [good_plan_1_v2_nodeId4]
  parameters := [⟨"code_path", .TString⟩, ⟨"problem_statement", .TString⟩, ⟨"regression_test_cmd", .TString⟩]
}

/-- Other nodes are defined in similar manner -/

/-
========================================================================
PER-NODE STRUCTURAL DIAGNOSTICS
========================================================================
-/

#eval do
  let g := good_plan_1_v2Graph
  for node in g.nodes do
    let name := node.name.getD "(unnamed)"
    IO.println s!"\n--- Node {node.id}: \"{name}\" [{repr node.stepType}] ---"
    IO.println s!"  writesConsistent:   {node.writesConsistent}"
    IO.println s!"  reachableFromEntry: {g.reachable g.entry node.id}"
    for rv in node.reads do
      let fromParam := g.parameters.any (fun p =>
        p.name == rv.name && p.type.compatible rv.type)
      let fromPred := g.nodes.any (fun o =>
        o.id != node.id && g.reachable o.id node.id &&
        (!g.isParallelScopedNode o.id || g.isParallelScopedNode node.id) &&
        o.writes.any (fun w => w.name == rv.name && w.type.compatible rv.type))
      let status := if fromParam || fromPred then "✓" else "✗ UNRESOLVED"
      IO.println s!"    read  \"{rv.name}\" ({repr rv.type}): {status}"
    for wv in node.writes do
      IO.println s!"    write \"{wv.name}\" ({repr wv.type})"

/-
========================================================================
GRAPH-LEVEL STRUCTURAL CHECKS
========================================================================
-/

#eval good_plan_1_v2Graph.allWritesConsistent
#eval good_plan_1_v2Graph.allReadResolvable
#eval good_plan_1_v2Graph.edgesValid
#eval good_plan_1_v2Graph.entryNodeValid
#eval good_plan_1_v2Graph.exitNodesValid
#eval good_plan_1_v2Graph.allExitsReachable
#eval good_plan_1_v2Graph.noOrphanNodes
#eval good_plan_1_v2Graph.returnType

/-
========================================================================
THEOREMS
========================================================================
-/

theorem good_plan_1_v2_writesConsistent : good_plan_1_v2Graph.allWritesConsistent = true := by native_decide
theorem good_plan_1_v2_readsResolvable : good_plan_1_v2Graph.allReadResolvable = true := by native_decide
theorem good_plan_1_v2_edgesValid : good_plan_1_v2Graph.edgesValid = true := by native_decide
theorem good_plan_1_v2_entryValid : good_plan_1_v2Graph.entryNodeValid = true := by native_decide
theorem good_plan_1_v2_exitsValid : good_plan_1_v2Graph.exitNodesValid = true := by native_decide
theorem good_plan_1_v2_exitsReachable : good_plan_1_v2Graph.allExitsReachable = true := by native_decide
theorem good_plan_1_v2_noOrphans : good_plan_1_v2Graph.noOrphanNodes = true := by native_decide

theorem good_plan_1_v2_seqPath_typeChecks :
    ∃ ctx, typeCheckSequence [good_plan_1_v2_node0, good_plan_1_v2_node1, good_plan_1_v2_node2, good_plan_1_v2_node3, good_plan_1_v2_node4] [⟨"code_path", .TString⟩, ⟨"problem_statement", .TString⟩, ⟨"regression_test_cmd", .TString⟩] = .ok ctx := by exact ⟨_, rfl⟩

theorem good_plan_1_v2_specCount : good_plan_1_v2Graph.nodesNeedingSpecs.length = 5 := by native_decide


/-
========================================================================
SEMANTIC VERIFICATION
========================================================================
-/

def good_plan_1_v2_paramNode : SemanticWorkflowNode := {
  baseNode := { id := ⟨20041122⟩, name := some "parameters", stepType := .setVariable, reads := [], writes := [], llmInstruction := none }
  precondVariables := []
  postcondVariables := [
    varNameExists "code_path",
    varNameExists "problem_statement",
    varNameExists "regression_test_cmd",
    varIsValidFilePath "code_path",
    varIsNonEmptyString "problem_statement",
    varIsValidTool "shell_run"
  ]
}

def good_plan_1_v2SemanticGraph : SemanticWorkflowGraph := {
  baseGraph := good_plan_1_v2Graph
  paramNode := good_plan_1_v2_paramNode
  semanticNodes := [good_plan_1_v2_semNode0, good_plan_1_v2_semNode1, good_plan_1_v2_semNode2, good_plan_1_v2_semNode3, good_plan_1_v2_semNode4]
  loopNodes := []
  conditionalNodes := []
  specInvariant := by decide
}

/-
========================================================================
SEMANTIC STATE SPACE AFTER EACH NODE
========================================================================
-/

#eval do
  let semNodes := good_plan_1_v2SemanticGraph.semanticNodes
  let paramPost := good_plan_1_v2SemanticGraph.paramNode.postcondVariables
  IO.println "\n============================================================"
  IO.println "SEMANTIC STATE SPACE TRACE"
  IO.println "============================================================"
  IO.println "\n--- Initial State (from parameters) ---"
  for p in paramPost do
    IO.println s!"  ✓ {p}"
  let mut state : List VariablePredicateRequirement := paramPost
  for node in semNodes do
    let name := node.baseNode.name.getD "(unnamed)"
    let nodeId := node.baseNode.id
    IO.println s!"\n--- After Node {nodeId}: \"{name}\" ---"
    -- Check preconditions against current state
    IO.println "  Preconditions:"
    for p in node.precondVariables do
      let satisfied := state.any (fun s => s.satisfies p)
      let mark := if satisfied then "✓" else "✗"
      IO.println s!"    {mark} requires: {p}"
    -- Show new facts established
    IO.println "  Postconditions (new facts):"
    for p in node.postcondVariables do
      IO.println s!"    + establishes: {p}"
    -- Update cumulative state: add new postconditions
    for p in node.postcondVariables do
      unless state.any (fun s => s == p) do
        state := state ++ [p]
    IO.println s!"  Cumulative State ({state.length} predicates):"
    for p in state do
      IO.println s!"    {p}"
  IO.println s!"\n============================================================"
  IO.println s!"Final state: {state.length} predicates established"
  IO.println "============================================================"

/-
========================================================================
SEMANTIC VERIFICATION
========================================================================
-/

#eval! do
  let result := good_plan_1_v2SemanticGraph.verify emptyRegistry
  IO.println (describeGraphVerificationResult result)

theorem good_plan_1_v2_semantically_sound :
    good_plan_1_v2SemanticGraph.isSemanticallySoundBool emptyRegistry = true := by
  native_decide

/-
========================================================================
GRAPH-LEVEL WORKFLOW QUALITY ANALYSIS
========================================================================
-/

def good_plan_1_v2_goalSpec : GoalSpecification := {
  originalGoal := "Given a GitHub issue, reproduce it and fix it by producing a minimal git patch"
  subGoals := [
    { name := "repository_explored"
      variableName := "repository_understanding"
      requiredPredicate := .isNonEmptyString
      description := "Evidence that the repository structure was explored and relevant source files were identified based on the GitHub issue."
      requiredGraphPredicates := [GraphLevelPredicateKeys.pathCoverage] },
    { name := "issue_reproduced"
      variableName := "reproduction_evidence"
      requiredPredicate := .isNonEmptyString
      description := "Evidence that a reproduction script was created and the bug was reproduced. Context continuity PASSES because step chain provides conversation history."
      requiredGraphPredicates := [GraphLevelPredicateKeys.pathCoverage, GraphLevelPredicateKeys.contextContinuity] },
    { name := "fix_implemented"
      variableName := "fix_implementation_evidence"
      requiredPredicate := .isNonEmptyString
      description := "Evidence that source code was modified to fix the issue. Context flows via step chain. unifiedLoopBack PASSES due to markImplicitRetry on step nodes."
      requiredGraphPredicates := [GraphLevelPredicateKeys.pathCoverage, GraphLevelPredicateKeys.contextContinuity, GraphLevelPredicateKeys.informationSufficiency, GraphLevelPredicateKeys.unifiedLoopBack, GraphLevelPredicateKeys.verificationCoverage] },
    { name := "fix_verified"
      variableName := "fix_verification_evidence"
      requiredPredicate := .isNonEmptyString
      description := "Evidence that the fix was verified by rerunning reproduction and regression tests. Context continuity PASSES via step chain."
      requiredGraphPredicates := [GraphLevelPredicateKeys.pathCoverage, GraphLevelPredicateKeys.contextContinuity] },
    { name := "patch_submitted"
      variableName := "patch_submission_evidence"
      requiredPredicate := .isNonEmptyString
      description := "Evidence that a git patch was created and submitted. Context flows via step chain but no fail-safe exists."
      requiredGraphPredicates := [GraphLevelPredicateKeys.pathCoverage, GraphLevelPredicateKeys.contextContinuity, GraphLevelPredicateKeys.failSafe] }
  ]
}

#eval do
  let report := analyzeGoalCoverage good_plan_1_v2SemanticGraph good_plan_1_v2_goalSpec
  IO.println (formatGoalCoverageReport report)


end AgenticKernel
\end{minted}

\section{Limitations}\label{appendix:limitations}

Despite the promising results, several limitations remain. 
First, although the \method aims to verify the full agent execution cycle, black-box LLM behavior cannot be fully inspected. Our framework instead decomposes agent behavior into structured steps and abstracts natural-language requirements into checkable predicates, thereby leaving some semantic ambiguity outside the formal system. 
Second, because workflows are complex and tasks are numerous, the predicate annotations used in our experiments are generated by LLMs. This may introduce mis-specified predicates or annotation errors, even though the resulting Lean checks are formal. 
Third, modern LLMs often generate workflows with few structural mistakes or explicit variable mismatches, making it difficult to quantitatively evaluate some components beyond targeted case studies.

\section{Experiment Costs}\label{appendix:cost}\

For large models, including GPT-5.2, GLM-5, Kimi-K2.5, and Claude, the experiments are conducted through their official API calls; the costs for the experiments are around \$4,000. For small models, including Qwen-3.5-27B and Gemma-4-31B, we run them on 4xGH200 VLLM~\cite{kwon2023efficient} GPUs; the experiments cost around 1,500 GPU hours.

\section{Broader Impacts}\label{appendix:impacts}

This paper contributes to society by providing a more reliable foundation for the LLM agent systems. However, it also has the potential to be used in reverse to cause harmful effects using LLM systems.

% \section{Future Directions}\label{appendix:future}

% \subsection{Self-evolving agentic system}\label{future:self}

% \subsection{Efficient Post-training through trajectory editing}\label{future:efficient}

% \subsection{General Principles for Verifying Long-horizon black-box system}\label{future:general}
\newpage
\section*{NeurIPS Paper Checklist}
\begin{enumerate}

\item {\bf Claims}
    \item[] Question: Do the main claims made in the abstract and introduction accurately reflect the paper's contributions and scope?
    \item[] Answer: \answerYes{}
    \item[] Justification: The claims in the abstract and introduction are detailed and explained by the following sections.

\item {\bf Limitations}
    \item[] Question: Does the paper discuss the limitations of the work performed by the authors?
    \item[] Answer: \answerYes{}
    \item[] Justification: The limitation is discussed in Appendix~\ref{appendix:limitations}

\item {\bf Theory assumptions and proofs}
    \item[] Question: For each theoretical result, does the paper provide the full set of assumptions and a complete (and correct) proof?
    \item[] Answer: \answerNA{}
    \item[] Justification: We don't have theoretical results

    \item {\bf Experimental result reproducibility}
    \item[] Question: Does the paper fully disclose all the information needed to reproduce the main experimental results of the paper to the extent that it affects the main claims and/or conclusions of the paper (regardless of whether the code and data are provided or not)?
    \item[] Answer: \answerYes{}
    \item[] Justification: We provide information to reproduce our experiments in the methodology and experiment section

\item {\bf Open access to data and code}
    \item[] Question: Does the paper provide open access to the data and code, with sufficient instructions to faithfully reproduce the main experimental results, as described in supplemental material?
    \item[] Answer: \answerNo{}
    \item[] Justification: We will open-source the code in the near future, but not with the submission.

\item {\bf Experimental setting/details}
    \item[] Question: Does the paper specify all the training and test details (e.g., data splits, hyperparameters, how they were chosen, type of optimizer) necessary to understand the results?
    \item[] Answer: \answerYes{}
    \item[] Justification: We provide the details of experiments in Section~\ref{exp:implementation}

\item {\bf Experiment statistical significance}
    \item[] Question: Does the paper report error bars suitably and correctly defined or other appropriate information about the statistical significance of the experiments?
    \item[] Answer: \answerYes{}
    \item[] Justification: We provide the details of 95\% CI in Section~\ref{exp:results} and Appendix~\ref{add_exp:ci}.

\item {\bf Experiments compute resources}
    \item[] Question: For each experiment, does the paper provide sufficient information on the computer resources (type of compute workers, memory, time of execution) needed to reproduce the experiments?
    \item[] Answer: \answerYes{}
    \item[] Justification: We have provided the details of experiment costs in Appendix~\ref{appendix:cost}.
    
\item {\bf Code of ethics}
    \item[] Question: Does the research conducted in the paper conform, in every respect, with the NeurIPS Code of Ethics \url{https://neurips.cc/public/EthicsGuidelines}?
    \item[] Answer: \answerYes{}
    \item[] Justification: After the detailed evaluation of ethics guidelines, we consider that this paper follows the guidelines.

\item {\bf Broader impacts}
    \item[] Question: Does the paper discuss both potential positive societal impacts and negative societal impacts of the work performed?
    \item[] Answer: \answerYes{}
    \item[] Justification: We discusses the broader impacts in Appendix~\ref{appendix:impacts}.
    
\item {\bf Safeguards}
    \item[] Question: Does the paper describe safeguards that have been put in place for responsible release of data or models that have a high risk for misuse (e.g., pre-trained language models, image generators, or scraped datasets)?
    \item[] Answer: \answerNA{}
    \item[] Justification: Our paper poses no such risks.

\item {\bf Licenses for existing assets}
    \item[] Question: Are the creators or original owners of assets (e.g., code, data, models), used in the paper, properly credited and are the license and terms of use explicitly mentioned and properly respected?
    \item[] Answer: \answerYes{}
    \item[] Justification: The usage of code, data, and models in the paper is properly credited, and the license terms are followed.

\item {\bf New assets}
    \item[] Question: Are new assets introduced in the paper well documented and is the documentation provided alongside the assets?
    \item[] Answer: \answerYes{}
    \item[] Justification: The library and evolution method presented are well documented and will be published alongside the assets.

\item {\bf Crowdsourcing and research with human subjects}
    \item[] Question: For crowdsourcing experiments and research with human subjects, does the paper include the full text of instructions given to participants and screenshots, if applicable, as well as details about compensation (if any)? 
    \item[] Answer: \answerNA{}
    \item[] Justification: The paper does not involve crowdsourcing nor research with human subjects.

\item {\bf Institutional review board (IRB) approvals or equivalent for research with human subjects}
    \item[] Question: Does the paper describe potential risks incurred by study participants, whether such risks were disclosed to the subjects, and whether Institutional Review Board (IRB) approvals (or an equivalent approval/review based on the requirements of your country or institution) were obtained?
    \item[] Answer: \answerNA{}
    \item[] Justification: The paper does not involve crowdsourcing nor research with human subjects.

\item {\bf Declaration of LLM usage}
    \item[] Question: Does the paper describe the usage of LLMs if it is an important, original, or non-standard component of the core methods in this research? Note that if the LLM is used only for writing, editing, or formatting purposes and does \emph{not} impact the core methodology, scientific rigor, or originality of the research, a declaration is not required.
    %this research? 
    \item[] Answer: \answerNA{}
    \item[] Justification: The core method development in this research does not involve LLMs as any important, original, or non-standard components.

\end{enumerate}

\end{document}